\documentclass[11pt,a4paper]{article}
\usepackage[margin=2.15cm]{geometry}
\usepackage{amsmath,amssymb,amsthm,booktabs,tabularx,longtable,array}
\usepackage{graphicx,xcolor,float,caption,enumitem,placeins}
\usepackage[numbers,sort&compress]{natbib}
\usepackage[hidelinks]{hyperref}
\hypersetup{
  pdftitle={Selective Hypergraph Refinement for Frozen Graph Clustering},
  pdfauthor={Zimo Si},
  pdfsubject={Label-free post-processing for frozen graph clustering},
  pdfkeywords={graph clustering, hypergraph learning, frozen model, label-free post-processing, selective refinement}
}
\setlist{nosep,leftmargin=1.8em}
\newcolumntype{Y}{>{\raggedright\arraybackslash}X}
\newcommand{\pp}{\,\mathrm{pp}}
\newcommand{\method}{SHR}
\theoremstyle{definition}
\newtheorem{definition}{Definition}[section]

\title{Selective Hypergraph Refinement for Frozen Graph Clustering}
\author{Zimo Si\\
\small Department of Mathematics, Faculty of Science, University of Macau\\
\small Macao SAR, China\\
\small Email: \texttt{uc32772@umac.mo}}
\date{}

\begin{document}
\maketitle

\begin{abstract}
Existing graph-clustering methods typically improve clustering performance by optimizing model parameters and node representations. Effective means of further improving the clustering results of an already trained and frozen model, however, remain limited. We study post-processing for frozen graph clustering. After checkpoint fixation, the procedure uses no labels and updates neither model parameters, node representations nor the original graph structure. Instead, it exploits an attribute hypergraph to supplement higher-order relations that ordinary graphs cannot readily express, thereby refining existing cluster assignments. Because global hypergraph refinement can yield both performance gains and erroneous updates, we propose Selective Hypergraph Refinement (\method). The method generates candidate residual directions from the hypergraph and evaluates their reliability using graph structure, node attributes and matched-null evidence. It updates only nodes with sufficient support and otherwise retains their original assignments. Further analysis shows that whether a node changes cluster is jointly governed by its native assignment gap and the directional strength of the refinement. In a controlled common-suite evaluation, 13 of 15 backbone--dataset cells had a positive mean macro gain, one produced exact no-action, and one was negative. The cell-equal macro gain was $0.066\pp$ (95\% bootstrap CI, $[0.030,0.107]\pp$), while only $0.209\%$ of hard assignments changed on average. A broader 15-combination native-interface evaluation yielded a macro gain of $0.137\pp$ at a mean change ratio of $0.375\%$. These results indicate that frozen clustering outputs retain a limited but measurable refinement space after training. The effect is heterogeneous across backbone--dataset pairs, and broader coverage also increases exposure to negative transfer.

\end{abstract}

\section{Introduction}

Existing graph-clustering methods primarily improve clustering performance by optimizing model parameters, node representations or graph structure\citep{zhang2025dese,zhu2026halo,xie2025dgac,ding2026sync,liu2026dgm,wang2026dbcd,xie2025onom}. However, further improving the clustering output of an already trained and frozen model remains challenging. We therefore study label-free post-processing for frozen graph clustering: whether existing cluster assignments can be improved without updating model parameters, node representations or the original graph structure.

Ordinary graphs primarily describe pairwise relations between nodes, whereas attribute hypergraphs can supplement higher-order associations induced by attribute similarity. For a frozen model, such information can no longer be incorporated through representation learning and can only be used to refine the existing clustering output. Our experiments reveal further scope for improvement through hypergraph refinement, but not every modification is beneficial. Broader refinement coverage can reach more potential errors while also disrupting assignments that were already correct. The key issue is therefore which candidate refinements should be accepted and how far the intervention should extend.

Motivated by this observation, we propose Selective Hypergraph Refinement (\method). \method{} uses an attribute hypergraph to generate candidate residual directions in the native cluster-coordinate system. It then evaluates different refinement states without labels by combining graph structure, node attributes and matched-null evidence. Only sufficiently supported modifications are retained; otherwise, the original assignments remain unchanged. We further find that, under a common global refinement strength, different nodes change their cluster assignments at different strength levels. We therefore define the critical refinement strength $\eta_{\mathrm{crit}}$ as the minimum strength required for a node to change its current cluster assignment for the first time. We evaluate \method{} across multiple graph-clustering backbones and datasets. The results show a positive aggregate gain under limited refinement coverage, together with pronounced backbone--dataset heterogeneity. Repair--harm analysis further shows that broader coverage creates more opportunities for repair but also greater exposure to negative transfer. Existing label-free signals, however, do not yet distinguish beneficial from harmful refinements consistently.

The main contributions of this work are as follows:
\begin{enumerate}
  \item \textbf{We formulate label-free post-processing for frozen graph clustering.} After checkpoint fixation, model parameters, node representations and the original graph structure remain unchanged, and only the existing cluster assignments are refined.
  \item \textbf{We propose Selective Hypergraph Refinement (\method).} The method uses an attribute hypergraph to generate candidate refinements and evaluates refinement states without labels through graph structure, node attributes and matched-null evidence. It retains the original result when the evidence is insufficient.
  \item \textbf{We characterize the relationship between refinement coverage and gain--risk behavior.} Critical-strength analysis explains implicit node selection under global refinement. Cross-backbone repair--harm experiments further show that broader coverage provides more opportunities for repair while increasing the risk of negative transfer.
\end{enumerate}

\section{Related Work}

\subsection{Graph Clustering and Hypergraph Learning}
Existing deep graph-clustering methods typically improve clustering performance by jointly optimizing node representations, graph structure or clustering objectives\citep{zhang2025dese,zhu2026halo,xie2025dgac,ding2026sync,liu2026dgm,wang2026dbcd,xie2025onom}. Their effectiveness depends substantially on continued updates to model parameters and node representations during training. Whereas an edge in an ordinary graph links only two nodes, a hyperedge can connect multiple nodes and thereby represent higher-order associations\citep{zhou2006hypergraphs}. Hypergraphs have therefore been increasingly used in graph clustering and representation learning. HCN learns clustering representations through hypergraph smoothing and structural alignment\citep{wang2025hcn}. HCN-PAI imputes missing attributes through higher-order propagation and further optimizes clustering results\citep{wang2025hcnpai}. JKHR jointly learns a consensus kernel and hypergraph regularization to integrate higher-order structural information\citep{niu2025jkhr}. Joint graph--hypergraph representations, higher-order encodings and multi-channel hypergraph models have also been used to integrate different forms of relational information\citep{peng2026tscgc,pellegrin2025encodings,quadir2026hgmn,chen2025dtshn}. These studies show that higher-order relations can complement ordinary graph structure, but most require node representations, graph structure, or the clustering space to be relearned during training. By contrast, we consider a graph-clustering backbone that has already been trained and remains frozen. The attribute hypergraph does not participate in new representation learning and serves only as auxiliary information for refining existing cluster assignments.

\subsection{Post-training Adaptation}
For an already trained graph model, test-time adaptation offers a means of improving performance without repeating the entire training process. Existing graph test-time adaptation methods typically use unlabeled test data to continue adjusting local model parameters, node representations or propagation mechanisms. For example, ASSESS adapts a pretrained model at test time through adaptive subgraph selection and prototype supervision\citep{zhao2025assess}. Other methods address structural shift by adjusting representations on the test graph through self-supervised low-rank feature tuning\citep{zhang2025avoiding}. Although TOTF adopts a train-once-then-freeze encoder design, it still extracts shared cross-view information and trains an auxiliary graph network after freezing to complete clustering\citep{li2025totf}. These approaches reduce the need to retrain the complete model, but still alter model representations, propagation processes or decision boundaries during testing or downstream clustering. \method{} imposes stricter constraints: after checkpoint fixation, the backbone, node embeddings and original graph structure remain unchanged, and all subsequent operations act only on the existing cluster assignments.

\subsection{Relational Post-processing}
Another line of work does not continue adapting the base model, but instead uses additional relational information to post-process existing predictions. After fixing the base predictions, Correct and Smooth refines semi-supervised node-classification results through residual propagation from labeled nodes and prediction smoothing\citep{huang2021correctsmooth}. NLCS (Nonlinear Correct and Smooth) further introduces non-linearity and higher-order relational modeling, allowing residuals to propagate along more complex node relations\citep{shao2025nlcs}. Graph structure has also been used in other forms of prediction post-processing. HGPF targets pretrained heterogeneous GNNs and adds an auxiliary relational-reasoning module after base-model training to improve classification\citep{yang2023hgpf}. These studies show that additional relational information can further refine the predictions of a base model. They nevertheless focus primarily on supervised or semi-supervised node classification, and some still depend on labels or post-training parameter optimization. \method{} instead targets a frozen model that has already completed clustering. It uses no labels after checkpoint fixation and does not create a new clustering space. Rather, an auxiliary hypergraph selectively refines existing cluster assignments, while the original result is retained when the evidence is insufficient.

\subsection{Uncertainty Modeling and State Averaging}
When several candidate models or clustering states are plausible, selecting a single optimum ignores uncertainty arising from model selection. Bayesian model averaging provides a classical means of retaining this uncertainty by weighting multiple candidate models\citep{hoeting1999bma}. Related ideas have also been applied to clustering. Cluster-ensemble methods integrate multiple base clusterings through probabilistic or structured models\citep{zhou2022ensemble}. Dynamic anchor-based methods using hypergraph reconstruction further construct and optimize a consensus structure from base clusterings\citep{xu2025dynamic}. clusterBMA weights multiple unsupervised clustering results using approximate posterior model probabilities, yielding a probabilistic final cluster assignment\citep{forbes2023clusterbma}. In parallel, research on safe multi-view clustering has shown that adding a view does not necessarily improve clustering performance and explicitly seeks to prevent performance degradation caused by additional views\citep{tang2022safe}. This concern is closely related to the gain--risk issue considered here, although these methods still learn new multi-view clustering representations through end-to-end training.

\method{} draws on the principle of retaining uncertainty among several plausible candidates rather than constructing a new Bayesian clustering model. For the frozen native soft cluster-assignment matrix $Q_0$, different refinement states represent different refinement coverages and strengths. \method{} assigns relative weights according to the label-free evidence and coverage complexity of each state, while jointly including the no-action state that leaves $Q_0$ unchanged. Unlike the Bayesian clustering methods above, which integrate multiple clustering results or re-estimate a consensus clustering, \method{} always refines $Q_0$ within the native cluster-coordinate system and creates no new clustering space.

\section{Preliminaries}
\label{sec:preliminaries}

\begin{definition}[Unsupervised Graph Clustering]
Consider an attributed undirected graph
$\mathcal{G}=(V,E,X)$,
where
$V=\{v_1,\ldots,v_N\}$
is the node set and
$E$ is the edge set,
while
$X\in\mathbb{R}^{N\times F}$
is the node-attribute matrix and
$A\in\{0,1\}^{N\times N}$
is the corresponding adjacency matrix.
Unsupervised graph clustering aims to partition the $N$ nodes into $K$ clusters based on graph structure and node attributes, without using node labels.

For an already trained graph-clustering model, denote its frozen node representation by
\begin{equation}
Z_0\in\mathbb{R}^{N\times d},
\end{equation}
and its soft cluster-assignment matrix by
\begin{equation}
Q_0\in[0,1]^{N\times K},
\qquad
\sum_{k=1}^{K}Q_{0,ik}=1.
\label{eq:prelim_q0}
\end{equation}
Here,
$Q_{0,ik}$
is the soft assignment of node $v_i$ to cluster $k$.
The current hard cluster assignment of node $v_i$ is
\begin{equation}
a_i=\arg\max_{k}Q_{0,ik}.
\label{eq:prelim_ai}
\end{equation}
\end{definition}

\begin{definition}[Attribute Hypergraph]
Unlike an edge in an ordinary graph, which connects only two nodes,
a hyperedge can connect multiple nodes simultaneously.
We define the attribute hypergraph as
\begin{equation}
\mathcal{H}=(V,\mathcal{E}_h),
\end{equation}
where
$\mathcal{E}_h=\{e_1,\ldots,e_M\}$
is the hyperedge set.
Its node--hyperedge incidence matrix is
\begin{equation}
B\in\{0,1\}^{N\times M},
\qquad
B_{ij}=
\begin{cases}
1, & v_i\in e_j,\\
0, & \text{otherwise}.
\end{cases}
\label{eq:hyper_incidence}
\end{equation}

Following the standard normalized form of hypergraph propagation\citep{zhou2006hypergraphs},
let $W$ denote the hyperedge-weight matrix,
and let $D_v$ and $D_e$ denote the node-degree and hyperedge-degree matrices, respectively.
The hypergraph propagation operator is
\begin{equation}
P_h=
D_v^{-1/2}
BWD_e^{-1}
B^{\top}
D_v^{-1/2}.
\label{eq:hyper_operator}
\end{equation}

The node and hyperedge degrees are respectively given by
\[
d(v_i)=\sum_{j=1}^{M}W_{jj}B_{ij},
\qquad
\delta(e_j)=\sum_{i=1}^{N}B_{ij}
\]
with
$D_v=\operatorname{diag}(d(v_1),\ldots,d(v_N))$
and
$D_e=\operatorname{diag}(\delta(e_1),\ldots,\delta(e_M))$.
Our implementation assigns equal weight to all hyperedges, and hence $W=I$.

We construct the auxiliary hypergraph from node attributes
and use $P_h$ to aggregate higher-order relations in the attribute space.
This hypergraph is used only as auxiliary information for refining the frozen backbone output
and neither replaces nor modifies the original graph structure.
\end{definition}

\begin{definition}[Frozen Graph-Clustering Refinement]
\label{def:frozen_refinement}
We consider an already trained and frozen graph-clustering model.
The available quantities are the original adjacency matrix $A$,
node attributes $X$,
the frozen node representation $Z_0$,
the native soft cluster-assignment matrix $Q_0$
and the number of clusters $K$.
The goal is to obtain a refined cluster-assignment matrix
without using node labels after checkpoint fixation
or updating model parameters, node representations, or the original graph structure:
\begin{equation}
Q^{*}
=
\mathcal{R}(A,X,Z_0,Q_0,K),
\qquad
Q^{*}\in[0,1]^{N\times K}.
\label{eq:frozen_refinement}
\end{equation}

The refinement acts only on the existing cluster assignments;
therefore, $Q^{*}$ and $Q_0$
share the same $K$-dimensional cluster-coordinate system.
When label-free evidence is insufficient, SHR simply retains the original result:
\begin{equation}
Q^{*}=Q_0.
\label{eq:no_action_prelim}
\end{equation}
\end{definition}

\begin{table}[H]
\centering
\caption{Principal notation and definitions used in this work.}
\label{tab:main_notations}
\small
\begin{tabularx}{\linewidth}{@{}lX@{}}
\toprule
Symbol & Definition \\
\midrule
$\mathcal{G}=(V,E,X)$ & Input graph with node attributes \\
$N$ & Number of nodes \\
$F$ & Dimensionality of node attributes \\
$A$ & Adjacency matrix of the original graph \\
$X$ & Node-attribute matrix \\
$K$ & Number of clusters \\
$Z_0$ & Node representation output by the frozen graph-clustering backbone \\
$Q_0$ & Native soft cluster-assignment matrix output by the frozen backbone \\
$a_i$ & Current hard cluster assignment of node $v_i$ under $Q_0$ \\
$\mathcal{H}$ & Auxiliary hypergraph constructed from node attributes \\
$\mathcal{E}_h$ & Hyperedge set \\
$B$ & Node--hyperedge incidence matrix \\
$P_h$ & Normalized hypergraph propagation operator \\
$\Delta$ & Candidate refinement residual mapped into the native cluster-coordinate system and formally defined in Section~\ref{sec:candidate_refinement} \\
$\eta$ & Candidate refinement strength \\
$Q^{*}$ & Refined soft cluster-assignment matrix output by \method{} \\
\bottomrule
\end{tabularx}
\end{table}
\FloatBarrier

\section{Methods}
\label{sec:methodology}

\subsection{Method Overview}

Under the setting in Section 3, SHR takes the original graph $A$, node attributes $X$, frozen representation $Z_0$, native soft assignments $Q_0$, and the number of clusters $K$ as input. It returns refined assignments $Q^\ast$ in the same cluster-coordinate system as $Q_0$. The backbone parameters, $Z_0$, and the original graph structure remain fixed throughout the procedure.

\begin{figure}[t]
    \centering
    \includegraphics[width=\textwidth]
    {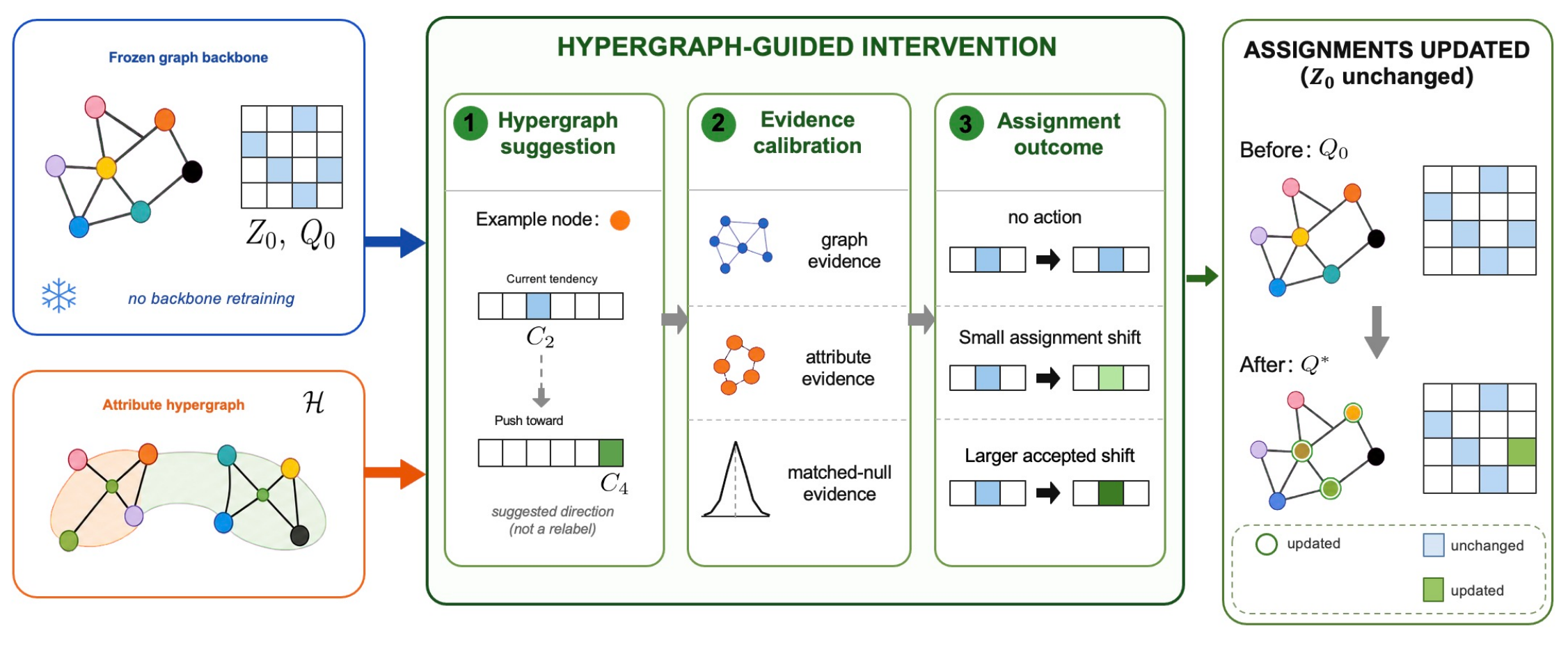}
    \caption{Overview of SHR. The frozen backbone provides $Z_0$ and $Q_0$, and the attribute hypergraph generates a candidate residual. Graph structure, node attributes, and matched-null evidence are used to evaluate the candidate states. The method ultimately updates only a subset of cluster assignments, while the model parameters, $Z_0$, and original graph structure remain frozen.}
    \label{fig:method}
\end{figure}

Figure~\ref{fig:method} separates candidate proposal from selective execution. The attribute hypergraph is used to construct a structured candidate residual $\Delta$ in the native cluster-logit coordinates. The proposal defines the available direction of change but does not by itself alter any assignment. SHR evaluates the proposal at the state, node, and run levels. At the state level, topology and attribute evidence, matched-null calibration, and coverage complexity assign relative support to different intervention extents. At the node level, a lower confidence bound removes changes whose support is unstable across repeated evidence estimates. At the run level, an attribute-provenance veto returns the complete output to $Q_0$ when the retained changes lack aggregate attribute support. Global shrinkage, change-rate control, and cluster-survival constraints limit the extent of the final update. SHR therefore uses label-free evidence to determine which hypergraph-generated changes are retained. The final update remains an additive correction in the native clustering logits and requires neither backbone retraining nor a transport plan.

\subsection{Hypergraph-Guided Candidate Proposal}
\label{sec:candidate_refinement}

\noindent\textbf{Attribute-hypergraph context.}
The original attributes $X$ are typically high-dimensional and sparse. SHR first applies truncated SVD and per-dimension standardization to obtain a low-rank attribute representation $Z_x$, and then searches for neighbors by cosine similarity. Each node and its $k_h=10$ attribute neighbors form a candidate hyperedge, and duplicate hyperedges are removed. Each hyperedge therefore represents a group of nodes that are similar in the attribute space. Its contribution is examined in the source ablation in Section~\ref{sec:ablation}.

Using the hypergraph propagation operator $P_h$ defined in Section~\ref{sec:preliminaries}, the node attributes and the outputs after one and two propagation steps are concatenated as
\begin{equation}
Z_h=
\operatorname{Std}
\left(
[Z_x\Vert P_hZ_x\Vert P_h^2Z_x]
\right).
\label{eq:multiscale_hyper}
\end{equation}
Here, $\operatorname{Std}(\cdot)$ denotes column-wise standardization and $\Vert$ denotes feature concatenation. $Z_x$ retains the node's own attributes, while $P_hZ_x$ and $P_h^2Z_x$ introduce higher-order context over different ranges. The concatenated representation is standardized to prevent one propagation scale from dominating the subsequent projection.

To extract cluster-level structure from $Z_h$, SHR applies a fixed, label-free $K$-means configuration with $K$ clusters. Let $\mu_k$ denote the $k$th cluster center and $D_{i,k}$ the squared distance from node $v_i$ to that center. A common softening scale $s$ is defined by the median, across nodes, of the distance to the nearest center:
\begin{equation}
\begin{aligned}
D_{i,k}
&=
\lVert Z_{h,i}-\mu_k\rVert_2^2,\\
s
&=
\max\left\{
\operatorname*{median}_{i}\min_kD_{i,k},
10^{-6}
\right\}.
\end{aligned}
\label{eq:aux_cluster_distance}
\end{equation}
The distances are then converted into auxiliary soft assignments:
\begin{equation}
\widetilde Q_{h,i,k}
=
\frac{
\exp\left[-(D_{i,k}-\min_jD_{i,j})/s\right]
}{
\sum_{\ell=1}^{K}
\exp\left[-(D_{i,\ell}-\min_jD_{i,j})/s\right]
}.
\label{eq:aux_soft_assignment}
\end{equation}
$\widetilde Q_{h,i,k}$ gives the soft membership of node $v_i$ in the $k$th auxiliary $K$-means cluster. Its columns have not yet been aligned with the semantics of $Q_0$.

Because the cluster indices produced by $K$-means are permutation-invariant, let $\mathfrak S_K$ denote all permutations of the $K$ cluster indices. SHR obtains a column permutation by maximizing the soft overlap between $\widetilde Q_h$ and $Q_0$:
\begin{equation}
\begin{aligned}
\pi^\ast
&=
\arg\max_{\pi\in\mathfrak S_K}
\sum_{i=1}^{N}\sum_{k=1}^{K}
\widetilde Q_{h,i,k}Q_{0,i,\pi(k)},\\
Q_{h,i,\pi^\ast(k)}
&=
\widetilde Q_{h,i,k}.
\end{aligned}
\label{eq:aux_column_alignment}
\end{equation}
This Hungarian matching removes only the cluster-ID permutation and does not use node labels. After alignment, $Q_h$ provides a cluster-level summary of the hypergraph representation. We concatenate it with $Z_h$ as
\begin{equation}
H_h=[Z_h\Vert Q_h].
\label{eq:hyper_context}
\end{equation}
$Q_h$ contributes only to constructing the candidate residual and is not used as the final clustering output.

\noindent\textbf{Representation-space alignment.}
$H_h$ and the frozen representation $Z_0$ have no direct correspondence in either dimensionality or coordinate axes. They are first projected to the common dimension
\[
d_c=
\min\left\{
32,\operatorname{col}(Z_0),\operatorname{col}(H_h)
\right\},
\]
where $\operatorname{col}(\cdot)$ denotes the number of matrix columns. After per-dimension centering and scaling, the projected representations are denoted by the frozen anchor coordinates $H_0$ and the hypergraph coordinates $\widetilde H_h$, respectively.

We then solve the orthogonal Procrustes alignment
\begin{equation}
R^\ast
=
\arg\min_{R^\top R=I}
\left\|
\widetilde H_hR-H_0
\right\|_F^2,
\qquad
H_h^\parallel
=
\widetilde H_hR^\ast.
\label{eq:hyper_alignment}
\end{equation}
Here, $R$ is an orthogonal matrix, $I$ is the identity matrix of the corresponding dimension, and $\|\cdot\|_F$ denotes the Frobenius norm. The orthogonality constraint restricts the alignment to rotations and reflections, avoiding additional scaling or nonlinear transformations.

\noindent\textbf{Mapping to the native clustering coordinates.}
Attribute similarity need not agree with local relationships in the original graph. Let
\[
\widetilde A=\operatorname{RowNorm}(A+I),
\]
where $\operatorname{RowNorm}(\cdot)$ denotes row normalization and $I$ is the node-dimensional identity matrix. The aligned hypergraph context is combined with its neighborhood aggregation on the original graph in a fixed ratio:
\begin{equation}
C_h
=
\frac{1}{2}H_h^\parallel
+
\frac{1}{2}\widetilde A H_h^\parallel.
\label{eq:aligned_hyper_context}
\end{equation}
$C_h$ therefore combines higher-order attribute context with local graph information.

$C_h$ remains in a representation space, whereas the refinement must act on the $K$ cluster logits of $Q_0$. SHR therefore fits a ridge decoder between the frozen anchor coordinates $H_0$ and the logit representation $\log(Q_0+\epsilon)$:
\begin{equation}
W_d
=
(H_0^\top H_0+\lambda I)^{-1}
H_0^\top
\log(Q_0+\epsilon).
\label{eq:ridge_decoder}
\end{equation}
Here, $\lambda>0$ is the ridge regularization coefficient, $\epsilon>0$ is a numerical-stability constant that prevents $\log 0$, and $\log(\cdot)$ is applied elementwise. $W_d$ maps the frozen representation coordinates to the current cluster-logit space.

Passing $C_h$ through the same decoder yields
\begin{equation}
\Delta=C_hW_d.
\label{eq:refinement_residual}
\end{equation}
Here, $\Delta\in\mathbb R^{N\times K}$, and $\Delta_{i,k}$ is the candidate increment for node $v_i$ along the $k$th existing cluster logit. Applying the same decoder to $C_h$ expresses the auxiliary context in the native cluster-logit coordinates of $Q_0$.

The resulting $\Delta$ is the fixed candidate direction used in the following stages and remains expressed in the native cluster-logit coordinates. No assignment is changed at this point, and the hypergraph does not determine the final intervention. Sections~4.3--4.5 use $\Delta$ to define an ordered set of candidate states; label-free evidence then determines the supported intervention extent, the rows that retain the proposed change, and whether the complete run reverts to $Q_0$. The subsequent label-free stage therefore operates on a fixed set of candidate states rather than optimizing an unrestricted $N\times K$ assignment update. The SVD dimension, K-means configuration, numerical safeguards, and ridge parameters are specified in Appendix~A.2.

\subsection{Critical Strength and an Effect-Calibrated Candidate Path}
\label{sec:implicit_selection}

Given the candidate direction $\Delta$, a scalar $\eta\geq0$ controls the global refinement strength. The candidate soft assignment for node $v_i$ is
\begin{equation}
Q_i(\eta)
=
\operatorname{softmax}
\left(
\log(Q_{0,i}+\epsilon)+\eta\Delta_i
\right).
\label{eq:global_refinement}
\end{equation}
Here, $Q_{0,i}$ and $\Delta_i$ are the $i$th rows of $Q_0$ and $\Delta$, respectively. Setting $\eta=0$ recovers the original assignment; the softmax ensures nonnegative rows that sum to one. All nodes share the same $\eta$, but each node has its own direction $\Delta_i$.

Let $a_i=\arg\max_kQ_{0,i,k}$ denote the current assignment of node $v_i$. For any candidate cluster $b\neq a_i$, define
\begin{equation}
\begin{aligned}
m_{i,b}
&=
\log Q_{0,i,a_i}-\log Q_{0,i,b},\\
d_{i,b}
&=
\Delta_{i,b}-\Delta_{i,a_i},\\
\eta_{i\rightarrow b}
&=
\frac{m_{i,b}}{d_{i,b}},
\qquad d_{i,b}>0.
\end{aligned}
\label{eq:margin_advantage}
\end{equation}
$m_{i,b}$ is the native logit margin of the current cluster over cluster $b$, and $d_{i,b}$ is the residual push toward cluster $b$ relative to the current cluster. Equating the two refined logits gives the critical value $\eta_{i\rightarrow b}=m_{i,b}/d_{i,b}$. If $d_{i,b}\leq0$, cluster $b$ cannot reach the score of the current cluster as $\eta$ increases.

Define
\[
\mathcal B_i=
\{b\neq a_i:d_{i,b}>0\}
\]
as the set of candidate clusters that can catch up with the current cluster. The strength at which node $v_i$ first reaches any assignment boundary is
\begin{equation}
\eta_{\mathrm{crit}}(i)
=
\begin{cases}
\displaystyle
\min_{b\in\mathcal B_i}
\frac{m_{i,b}}{d_{i,b}},
&
\mathcal B_i\neq\varnothing,\\[3mm]
+\infty,
&
\mathcal B_i=\varnothing,
\end{cases}
\label{eq:eta_crit}
\end{equation}
When $\mathcal B_i\neq\varnothing$, the corresponding first target cluster is
\begin{equation}
b_{\mathrm{crit}}(i)
=
\arg\min_{b\in\mathcal B_i}
\frac{m_{i,b}}{d_{i,b}}.
\label{eq:b_crit}
\end{equation}

A smaller native margin or a stronger relative push lowers $\eta_{\mathrm{crit}}(i)$. Under a common global $\eta$, nodes with smaller critical strengths cross their assignment boundaries earlier. The boundary $\eta=\eta_{\mathrm{crit}}(i)$ corresponds to tied logits; a strict hard-assignment change occurs only after the boundary is crossed.

The margins of $Q_0$ and the residual scale differ across backbones, so the same numerical $\eta$ is not comparable across systems. SHR uses seven target hard-assignment change ratios---$0.5\%$, $1\%$, $2\%$, $3\%$, $5\%$, $8\%$, and $12\%$---and selects, from a frozen $\eta$ grid, the state closest to each target coverage. The grid and selection rule are specified in Appendix~\ref{app:frozen_constants}.

The critical strengths induce an ordering of candidate assignment changes along the fixed residual. $\eta_{\mathrm{crit}}$ determines when each node reaches an assignment boundary. The subsequent coverage calibration allows candidate states to be compared across backbones. It provides no evidence that the change is beneficial. The states $Q^{(\eta_r)}$ are therefore treated only as proposals and are compared with the unchanged state using label-free evidence in the next subsection.

\begin{figure}[t]
    \centering
    \includegraphics[width=\textwidth]
    {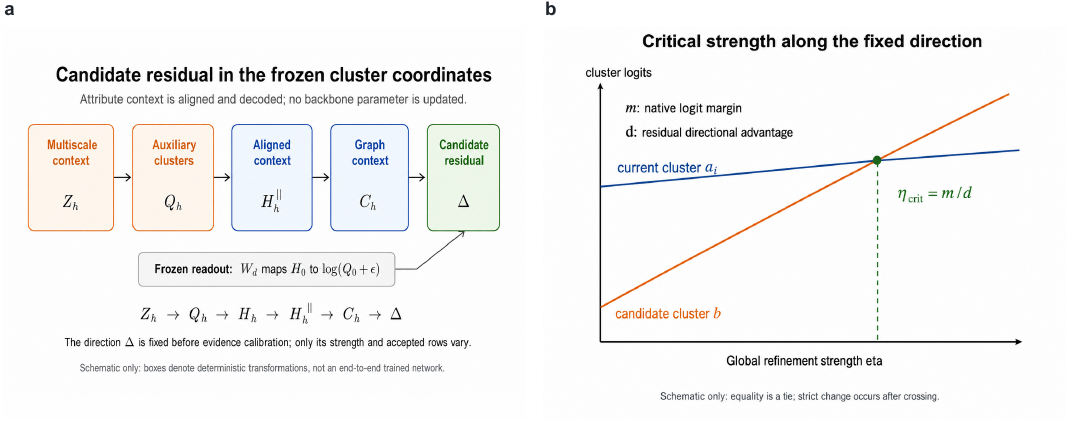}
    \caption{Candidate-residual construction and critical refinement strength. (a) The multiscale attribute context is summarized by auxiliary clusters, aligned with the frozen representation, combined with local graph context, and decoded through the frozen readout to obtain $\Delta$; no backbone parameter is updated. (b) After $\Delta$ is fixed, the logits of the current and candidate clusters vary with $\eta$ and intersect at $\eta_{\mathrm{crit}}$.}
    \label{fig:formula_mechanism}
\end{figure}

\subsection{Evidence-Driven State Selection}
\label{sec:evidence_assessment}

The effect-calibrated path from Section~4.3 contains seven nonzero candidate states that differ in intervention extent. Their order reflects how readily nodes respond to the fixed residual, not whether the resulting changes are beneficial. SHR therefore treats every $Q^{(\eta_r)}$ as a proposal and compares it with the unchanged output $Q^{(0)}=Q_0$ using label-free evidence. Let $\mathcal S$ denote the index set of the nonzero states. For $r\in\mathcal S$, define
\[
\mathcal C_r
=
\left\{
i:
\arg\max_kQ^{(\eta_r)}_{i,k}
\neq a_i
\right\},
\qquad
c_r=|\mathcal C_r|,
\]
where $\mathcal C_r$ is the set of nodes whose hard assignments change in state $r$, and $c_r=|\mathcal C_r|$ is its intervention size.

We evaluate each state using topology and attribute evidence. Topology evidence asks whether the candidate cluster is more compatible than the current cluster with the node's neighborhood pattern in the original graph. Attribute evidence measures the same candidate-versus-current compatibility in the low-rank attribute space. Both are candidate-versus-current scores estimated through held-out folds. The node utility also subtracts a Jensen--Shannon divergence penalty~\citep{lin1991js} to discourage large changes in the soft assignments. The exact predictive scores and utility are given in Appendix~A.1.

High internal support alone is not sufficient, because low-confidence nodes and broad states can receive favorable scores even when the node--residual correspondence is uninformative. SHR therefore constructs matched-null controls by permuting the rows of $\Delta$ within the same baseline cluster and confidence quintile, followed by recalibration to the same target coverage. The permutation approximately preserves node difficulty and intervention size while removing the original node--direction correspondence. The empirical odds $e_r$ summarize whether the observed correspondence receives more support than its matched controls. They are relative evidence scores, not probabilities of correctness or Bayes factors.

Broader states have more operational freedom and can obtain favorable aggregate support by modifying more nodes. SHR accounts for this difference through the coverage complexity $L(c_r)$, which combines a sparse count prior and an MDL penalty. The coverage term therefore penalizes states that modify more nodes. The unchanged output remains an explicit competing state rather than a fallback added after selection. The log weights are
\begin{align}
\ell_0
&=
L(0),
&
\ell_r
&=
\log e_r+L(c_r)-\log|\mathcal S|,
\quad r\in\mathcal S.
\label{eq:state_log_weights}
\end{align}
Here, $|\mathcal S|$ is the number of nonzero proposals, and the term $-\log|\mathcal S|$ controls the total weight introduced by considering several nonzero states.

For $j\in\{0\}\cup\mathcal S$, the normalized weights are defined as
\begin{equation}
\omega_j
=
\frac{\exp(\ell_j)}
{\exp(\ell_0)+
 \sum_{s\in\mathcal S}\exp(\ell_s)}.
\label{eq:state_weights}
\end{equation}
The mixed candidate is then
\begin{equation}
Q_{\mathrm{mix}}
=
\omega_0Q_0
+
\sum_{r\in\mathcal S}
\omega_rQ^{(\eta_r)}.
\label{eq:state_weighting}
\end{equation}
The normalized weights $\omega_j$ encode relative label-free support and are not posterior probabilities. Including $Q_0$ allows SHR to select no action when the nonzero proposals are weakly supported. State averaging also reduces abrupt coverage changes caused by small differences in evidence. $Q_{\mathrm{mix}}$ summarizes the state-level evidence and is passed to the node-level filtering stage rather than used directly as the final output. Section~4.5 applies node-level evidence filtering and the run-level no-action decision.

\subsection{Node- and Run-Level Selective Execution}
\label{sec:selective_execution}

State-level weighting determines the overall intervention extent, but individual node updates may still be unreliable. Before producing the final output, SHR first limits the global update and then filters the remaining node changes. It first constructs the fixed shrinkage path
\begin{equation}
Q_\tau
=
\operatorname{Normalize}
\left(
(1-\tau)Q_0+\tau Q_{\mathrm{mix}}
\right).
\label{eq:safety_projection}
\end{equation}
Here, $\tau\in[0,1]$ is the shrinkage coefficient, and $\operatorname{Normalize}(\cdot)$ denotes row-wise normalization. The path is evaluated from larger to smaller values of $\tau$. For each candidate, SHR checks the hard-assignment change ratio, mean Jensen--Shannon divergence, and number of active clusters. They constrain the magnitude and structural validity of the update. The first candidate that satisfies all three conditions is denoted by $Q_{\mathrm{safe}}$.

SHR then considers only nodes whose hard assignments change under $Q_{\mathrm{safe}}$. For node $v_i$, let $\bar u_i$ and $s_i$ denote the mean and standard deviation of its support across the repeated label-free evidence calculations. The node-level lower confidence bound is
\begin{equation}
\operatorname{LCB}_i
=
\bar u_i
-
1.645\frac{s_i}{\sqrt R}.
\label{eq:node_lcb}
\end{equation}
Here, $R=5$ is the number of evidence repetitions, and 1.645 is the one-sided 95\% normal quantile used by the frozen protocol. A positive LCB indicates that the node retains positive support after accounting for variation across evidence repeats. Nodes with a nonpositive LCB revert to $Q_{0,i}$. Cluster survival is applied afterward as a structural constraint so that no original cluster loses all of its nodes.

Let $\mathcal A_{\mathrm{acc}}$ denote the nodes that pass the LCB and cluster-survival checks. Finally, the retained nodes must also receive positive mean attribute support at the run level. If $\mathcal A_{\mathrm{acc}}$ is empty or its mean attribute support is nonpositive, the attribute-provenance veto returns the complete output to $Q_0$. Otherwise, only the accepted rows are retained:
\begin{equation}
Q_i^\ast
=
\begin{cases}
Q_{\mathrm{safe},i},
&
i\in\mathcal A_{\mathrm{acc}},\\
Q_{0,i},
&
i\notin\mathcal A_{\mathrm{acc}}.
\end{cases}
\label{eq:selective_row_rollback}
\end{equation}
If no node passes the filtering stage, or if the run-level attribute provenance check fails, the entire run returns
\[
Q^\ast=Q_0.
\]

Thus, state-level weighting determines the intervention extent, node-level LCB filtering retains supported rows, and the provenance check can still return the entire run to $Q_0$. Shrinkage and cluster-survival checks enforce feasibility. These controls restrict the intervention but do not guarantee an improvement in external clustering metrics. The fixed projection path, thresholds, and other execution constants are reported in Appendix~A.2.

\section{Experiments}
\label{sec:experiments}

We evaluate SHR from four perspectives: overall clustering performance, the contribution of its main components, the mechanism of selective refinement, and robustness across frozen backbones and datasets. The experiments address the following questions:

\noindent\textbf{Q1:} Can \method{} improve existing cluster assignments across different frozen graph-clustering backbones and datasets, and is its performance consistent across backbones?

\noindent\textbf{Q2:} Do the key components of \method{} and the higher-order information supplied by the attribute hypergraph contribute to the final refinement results?

\noindent\textbf{Q3:} How do refinement strength and refinement coverage affect node changes, clustering gains, and negative transfer, and how does the selective-refinement mechanism of \method{} operate?

\noindent\textbf{Q4:} Are the aggregate conclusions for \method{} stable across backbones, aggregation schemes, and boundary conditions, and what are its principal applicability limitations?

\subsection{Experimental Setup}
\label{sec:experiment_setup}

\noindent\textbf{Datasets and Backbones.}
We evaluated \method{} on five frozen graph-clustering backbones, DeSE, HALO, DGAC, SynC, and DGM, across attributed-graph datasets including Cora, Citeseer, Photo, ACM, UAT, EAT, BAT, DBLP, and Adam. Multiple backbones are included to test whether SHR can operate on different frozen clustering outputs. SHR reads the frozen node representations and native cluster assignments produced by each backbone without retraining the backbone. The broader native-interface evaluation comprises 15 backbone--dataset combinations with valid native cluster assignments. The frozen-embedding KMeans readout for DeSE--Computers and cases that do not satisfy the interface requirements are included only in supplementary analyses and are excluded from that aggregate. Table~\ref{tab:registry} summarizes the evaluation scope, and Appendix~\ref{app:provenance} provides dataset statistics, preprocessing details, and checkpoint provenance.

\noindent\textbf{Controlled and Broader Evaluation Protocols.}
We use two complementary evaluation protocols. First, we constructed a complete controlled common suite comprising HALO, DGAC, and SynC on Cora, Citeseer, ACM, UAT, and DBLP. Each backbone--dataset cell contains five paired frozen runs, giving 75 experimental units. All three backbones provide a valid native $Q_0$ on all five datasets; inclusion was determined by frozen-interface validity rather than SHR performance. Second, we retained a broader native-interface evaluation covering DeSE, HALO, DGAC, SynC, and DGM, with 15 valid backbone--dataset combinations and 113 frozen units. The broader evaluation covers more backbone-native datasets, although the dataset sets and frozen-unit counts differ across backbones. It is used to examine cross-system heterogeneity and robustness, while the controlled common suite provides the direct paired comparison. The preregistered $4\times5$ matrix and its DeSE interface failures are documented in Appendix~C.3.

\begin{table}[!htbp]
\centering
\caption{Backbones, datasets, and their evaluation scope. Broader combinations enter the 15-combination native-interface evaluation; Development and Boundary combinations are used only for supplementary analyses.}
\label{tab:registry}
\small
\begin{tabularx}{\textwidth}{l l l Y}
\toprule
Backbone & Dataset & Evaluation scope & Notes \\
\midrule
DeSE\citep{zhang2025dese} & Cora, Citeseer, Photo & Broader & Native assignment; label-free after checkpoint fixation \\
HALO\citep{zhu2026halo} & ACM, UAT, EAT, BAT & Broader & Broader native-interface evaluation \\
DGAC\citep{xie2025dgac} & Cora, Citeseer & Broader & Includes valid no-action \\
SynC\citep{ding2026sync} & Cora, Citeseer, ACM, UAT, DBLP & Broader & Includes negative cases \\
DGM\citep{liu2026dgm} & Adam & Broader & Native near-no-action case \\
DBCD\citep{wang2026dbcd} & Cora & Boundary & Interface boundary \\
DeSE strict-final & Photo, Computers & Boundary & Degenerated-anchor boundary \\
DeSE legacy & Computers & Development & Development-only non-native readout \\
\bottomrule
\end{tabularx}
\end{table}

\noindent\textbf{Comparison Methods.}
All comparison methods share the same frozen assignments $Q_0$ and candidate residual $\Delta$, allowing the effects of refinement strength, coverage, and node-level selection to be compared under a common proposal. Global-A is the simplest global baseline and applies a fixed refinement strength to every node. Strength-Bayes replaces this single strength with a weighted set of refinement-strength states to examine the effect of strength uncertainty. CS-BAYES further considers refinement coverage and strength jointly, whereas \method{} adds node-level selection after state assessment. CM-Global matches the changed-node count of CS-BAYES and is used only to diagnose the effect of broader coverage. The exact state spaces, priors, and computational rules are reported in Appendix~\ref{app:comparators} (Table~\ref{tab:comparators}).

\noindent\textbf{Evaluation Metrics and Protocol.}
We evaluated final clustering quality using ACC, NMI, ARI, and macro-F1; ACC and macro-F1 were computed after Hungarian matching. For each metric, the change relative to the frozen baseline is reported in percentage points. Macro gain is the equally weighted mean of the four metric gains, and Change is the proportion of nodes whose final hard assignments differ from the frozen baseline. In addition to the final clustering metrics, we used Repair Recall and Harm Fraction for node-level post hoc diagnosis. They describe, respectively, the extent to which repairable errors in the fixed candidate space are ultimately corrected and the proportion of changed nodes that move from baseline-correct to incorrect assignments. Both are post-hoc diagnostics computed after the SHR outputs are frozen; their definitions are given in Appendix~B. Common-suite cell-level summaries report the mean, SD, median, range, and a descriptive 95\% interval from $10{,}000$ paired seed-level bootstrap resamples. The aggregate common-suite estimator first forms 15 cell means and then uses the cells as sampling units in $10{,}000$ bootstrap resamples, with random seed $4102026$. The broader native-interface evaluation uses combination-equal aggregation: frozen runs are first aggregated within each backbone--dataset combination and the 15 combination-level results are then averaged equally. Backbone-equal, run-weighted, and leave-one-backbone-out summaries are used only for robustness assessment. To describe negative transfer, a single run with macro gain $g_r<0$ was defined as a negative run; a more pronounced negative-tail outcome with $g_r\leq-0.25\pp$ was defined as a catastrophic run; and a combination whose mean macro gain was below zero was defined as a negative combination. Runtime measurement begins after $A$, $X$, $Z_0$, and $Q_0$ have been prepared and ends when formal SHR returns $Q^*$. Each frozen unit receives one warm-up and three repetitions timed with \texttt{time.perf\_counter()}, whose median is reported; backbone training is excluded. The \method{} refinement stage was label-free after checkpoint fixation. The DeSE paper-best checkpoints were selected upstream using ground-truth labels, so the label-free claim does not extend to backbone training or checkpoint selection.

Protocol summary. The controlled common suite contains 75 frozen units, while the broader native-interface evaluation contains 113. Separate historical and mechanism registries are used only for component and post-hoc diagnostic analyses. Results from these registries are reported separately and are not pooled.

\FloatBarrier
\suppressfloats[t]
\subsection{Overall Performance}
\label{sec:overall_performance}

\noindent\textbf{Controlled common-suite performance.}
Table~3 reports the absolute ACC, NMI, ARI, and F1 before and after SHR for each frozen backbone. The comparison is paired within each backbone rather than across different backbone models.

\begin{table}[!htbp]
\centering
\caption{Controlled common-suite results before and after SHR. For each frozen backbone, the baseline and its SHR-refined output are reported consecutively on the same five datasets. Each entry is averaged over five paired frozen runs. ACC, NMI, ARI, and F1 are reported on the $[0,1]$ scale. Boldface indicates the better value within each backbone--SHR pair based on the unrounded means and is not used to rank different backbones.}
\label{tab:common_suite_paired}
\scriptsize
\setlength{\tabcolsep}{2.5pt}
\begin{tabular}{@{}l*{12}{r}@{}}
\toprule
\multicolumn{13}{l}{\textbf{(a) Cora, Citeseer, and ACM}} \\
 & \multicolumn{4}{c}{Cora} & \multicolumn{4}{c}{Citeseer} & \multicolumn{4}{c}{ACM} \\
\cmidrule(lr){2-5}\cmidrule(lr){6-9}\cmidrule(lr){10-13}
Method & ACC & NMI & ARI & F1 & ACC & NMI & ARI & F1 & ACC & NMI & ARI & F1 \\
\midrule
HALO & 0.5078 & 0.3853 & 0.2634 & \textbf{0.4403} & 0.4916 & 0.2597 & 0.2354 & 0.4415 & 0.8282 & 0.5334 & 0.5755 & 0.8262 \\
HALO + SHR & \textbf{0.5086} & \textbf{0.3865} & \textbf{0.2639} & 0.4399 & \textbf{0.4926} & \textbf{0.2603} & \textbf{0.2363} & \textbf{0.4424} & \textbf{0.8294} & \textbf{0.5354} & \textbf{0.5778} & \textbf{0.8275} \\
\addlinespace[2pt]
DGAC & 0.6393 & 0.5339 & 0.4381 & 0.6344 & 0.6909 & 0.4350 & 0.4450 & 0.6444 & 0.3647 & 0.0067 & 0.0051 & 0.2890 \\
DGAC + SHR & \textbf{0.6405} & \textbf{0.5351} & \textbf{0.4397} & \textbf{0.6350} & 0.6909 & 0.4350 & 0.4450 & 0.6444 & \textbf{0.3648} & \textbf{0.0068} & \textbf{0.0052} & \textbf{0.2891} \\
\addlinespace[2pt]
SynC & 0.6744 & 0.5034 & 0.4336 & 0.6676 & \textbf{0.6939} & 0.4426 & 0.4538 & \textbf{0.6410} & 0.9155 & 0.7036 & 0.7647 & 0.9157 \\
SynC + SHR & \textbf{0.6763} & \textbf{0.5058} & \textbf{0.4364} & \textbf{0.6694} & 0.6938 & \textbf{0.4428} & \textbf{0.4538} & 0.6410 & \textbf{0.9157} & \textbf{0.7040} & \textbf{0.7652} & \textbf{0.9159} \\
\bottomrule
\end{tabular}
\vspace{5pt}

\begin{tabular}{@{}l*{8}{r}@{}}
\toprule
\multicolumn{9}{l}{\textbf{(b) UAT and DBLP}} \\
 & \multicolumn{4}{c}{UAT} & \multicolumn{4}{c}{DBLP} \\
\cmidrule(lr){2-5}\cmidrule(lr){6-9}
Method & ACC & NMI & ARI & F1 & ACC & NMI & ARI & F1 \\
\midrule
HALO & 0.4951 & 0.2281 & 0.1933 & 0.4642 & 0.3062 & 0.0259 & -0.0010 & 0.1554 \\
HALO + SHR & \textbf{0.4966} & \textbf{0.2305} & \textbf{0.1955} & \textbf{0.4653} & \textbf{0.3062} & \textbf{0.0260} & \textbf{-0.0010} & \textbf{0.1555} \\
\addlinespace[2pt]
DGAC & 0.4963 & 0.2182 & 0.1961 & 0.4740 & 0.3202 & 0.0232 & 0.0213 & 0.2733 \\
DGAC + SHR & \textbf{0.4973} & \textbf{0.2189} & \textbf{0.1970} & \textbf{0.4753} & \textbf{0.3204} & \textbf{0.0232} & \textbf{0.0214} & \textbf{0.2734} \\
\addlinespace[2pt]
SynC & \textbf{0.5555} & \textbf{0.2584} & \textbf{0.2485} & \textbf{0.5478} & 0.8104 & 0.5140 & 0.5695 & 0.8049 \\
SynC + SHR & 0.5551 & 0.2579 & 0.2484 & 0.5471 & \textbf{0.8107} & \textbf{0.5150} & \textbf{0.5703} & \textbf{0.8052} \\
\bottomrule
\end{tabular}
\end{table}

The paired results show small positive changes in most backbone--dataset cells, but not improvement in every case. Thirteen of the 15 cells had a positive mean macro gain, DGAC--Citeseer retained exact no-action, and SynC--UAT was negative. Equally weighting the 15 cell means yielded a macro gain of $0.066\pp$ with a 95\% combination-bootstrap CI of $[0.030,0.107]\pp$; the cell median was $0.050\pp$, the IQR was $[0.006,0.105]\pp$, and the range was $[-0.042,0.224]\pp$. The common suite therefore shows a small overall positive effect, with clear variation across backbone--dataset pairs. Complete seed-level dispersion and cell-level bootstrap intervals are reported in Appendix~\ref{app:common_suite}.

\FloatBarrier
\noindent\textbf{Broader native-interface evaluation.}

\begin{figure}[!htbp]
\centering
\includegraphics[width=0.98\linewidth]{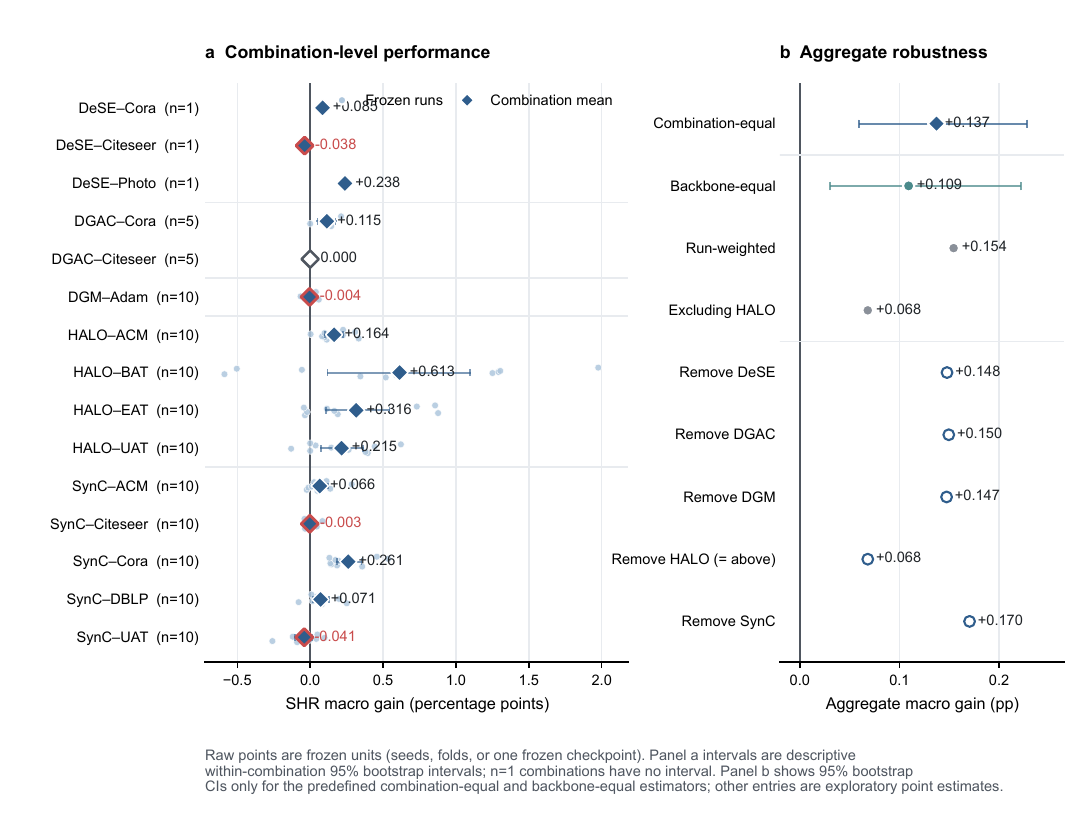}
\caption{Performance heterogeneity and robustness of SHR across backbone--dataset combinations. (a) Macro gain for each of the 15 combinations in the broader native-interface evaluation. Pale points denote frozen runs, highlighted markers denote combination means, and the vertical zero line denotes no change relative to the frozen baseline. Combinations with multiple frozen runs also show descriptive intervals; the three single-checkpoint DeSE combinations are shown as points only, without artificial intervals. (b) Aggregate macro gain under alternative aggregation schemes and leave-one-backbone-out analyses. Combination-equal aggregation is the main estimand for this broader evaluation; all other results are robustness or exploratory analyses. Frozen units differ by backbone and comprise seeds, folds, or one frozen checkpoint.}
\label{fig:shr_heterogeneity}
\end{figure}

We further evaluate 15 valid native-interface combinations spanning five backbones. Figure~\ref{fig:shr_heterogeneity}a shows that 10 combinations had positive macro gain, DGAC--Citeseer returned exact no-action, and four combinations were negative. With equal weighting across combinations\footnote{Combination-equal aggregation first summarizes frozen runs within each backbone--dataset combination and then assigns equal weight to all 15 combinations; it is the main statistical estimand for the broader evaluation. Backbone-equal aggregation first averages dataset-level results within each backbone and then weights the five backbones equally. Run-weighted aggregation assigns equal weight directly to all frozen runs, so combinations with more frozen runs contribute more to the overall result. Leave-one-backbone-out analysis removes one backbone at a time and recomputes the combination-equal result to assess whether the aggregate conclusion is dominated by a particular backbone.}, the mean ACC, NMI, ARI, and F1 gains were $0.103$, $0.171$, $0.152$, and $0.123\pp$, respectively, yielding a macro gain of $0.137\pp$ with a 95\% combination-bootstrap CI of $[0.059,0.228]\pp$. Because the dataset sets differ across backbones, this analysis is used to assess cross-system behavior rather than as a direct replacement for the common-suite comparison. Complete combination-level absolute baseline/final values and gains are provided in Appendix~\ref{app:complete_results}.

\FloatBarrier
\noindent\textbf{Robustness of broader evaluation.}
Figure~\ref{fig:shr_heterogeneity}b compares alternative aggregation rules and leave-one-backbone-out estimates. The combination-equal, backbone-equal, and run-weighted macro gains remained positive at $0.137$, $0.109$, and $0.154\pp$, respectively. Leave-one-backbone-out estimates ranged from $0.068$ to $0.170\pp$, with the lowest value obtained after removing HALO. Thus, HALO contributes substantially to the aggregate effect, although the mean remains positive without it.

\noindent\textbf{Intervention scale and post-processing cost.}
Performance changes should be interpreted together with their intervention scale. Table~\ref{tab:common_suite_intervention_runtime} shows that the equally weighted mean of the 15 cell-level change-ratio means was $0.209\%$; the median was $0.159\%$, the IQR was $[0.077\%,0.343\%]$, and the mean corresponded to 2.09 changed nodes per 1,000. Fourteen of the 15 cells had a mean change ratio below $0.5\%$, and all were below $1\%$. The broader 15-combination/113-run evaluation had a separate mean change ratio of $0.375\%$; these statistics describe different registries and are not pooled. The observed gains therefore arise from changes to only a small fraction of the frozen assignments.

SHR operates after checkpoint fixation and requires no backbone retraining. Under the current CPU/NumPy--SciPy implementation and tested graph scales, the median post-processing wall time from frozen inputs to $Q^*$ across the 75 units was $1.71$ s (IQR, $1.48$--$2.01$ s). Dataset-level medians ranged from $0.90$ to $2.09$ s. Backbone training time is excluded because it was not measured under the same hardware and protocol. Full intervention and timing details for all 15 cells are reported in Appendix~\ref{app:common_suite}.

\begin{table}[!htbp]
\centering
\caption{Intervention scale and post-processing cost on the common suite. Macro gain and Change first average the five seeds within a backbone--dataset cell and then weight the three backbones equally for each dataset. Runtime is the median across 15 frozen units per dataset after one warm-up and three timed repetitions; it excludes backbone training.}
\label{tab:common_suite_intervention_runtime}
\small
\begin{tabular}{lrrrr}
\toprule
Dataset & $N$ & Macro gain (pp) & Change (\%) & SHR time (s) \\
\midrule
Cora & 2708 & +0.129 & 0.438 & 1.71 \\
Citeseer & 3327 & +0.028 & 0.064 & 2.09 \\
ACM & 3025 & +0.070 & 0.132 & 1.50 \\
UAT & 1190 & +0.078 & 0.314 & 0.90 \\
DBLP & 4057 & +0.026 & 0.097 & 1.98 \\
Overall & -- & +0.066 & 0.209 & 1.71 \\
\bottomrule
\end{tabular}
\end{table}

\FloatBarrier
\subsection{Ablation Study}
\label{sec:ablation}
\suppressfloats[t]

We study two types of ablation: method components and auxiliary relation sources. Component ablations examine matched-null evidence, coverage control, state averaging, and node-level execution. Source ablations replace only the relation used to construct the candidate residual while keeping the frozen backbone and the downstream SHR procedure fixed.

\noindent\textbf{Component ablation.} We compare the complete method with several simplified variants. \textbf{Effect-only} inherits the changed-node count selected by complete SHR and therefore serves only as a matched-coverage diagnostic. \textbf{No matched-null} retains the label-free evidence from graph structure, node attributes, and assignment displacement, but removes the matched-null random control, so a real candidate state no longer has to outperform null states obtained after disrupting the node--residual correspondence. \textbf{No coverage/MDL} removes both the coverage prior based on changed-node count and the MDL complexity penalty, so states that modify more nodes are no longer additionally constrained for their greater freedom. \textbf{No node-level filtering + provenance veto} directly uses a candidate result after it passes the global magnitude constraints; it no longer applies Combined-LCB or cluster-survival screening to changed nodes and no longer rejects an entire run according to the mean attribute support of retained nodes. \textbf{MAP} retains the candidate states and their weight calculation, but replaces averaging over the no-action and multiple non-zero states with the single state having the largest weight.

\begin{table}[!htbp]
\centering
\small
\setlength{\tabcolsep}{5pt}
\caption{Component ablation of SHR. Results are from the frozen historical distillation registry and are used only to compare the performance--risk profiles of method simplifications; their absolute values are not mixed with those from the current broader native-interface evaluation.}
\label{tab:component_ablation}
\begin{tabularx}{0.99\linewidth}{@{}>{\raggedright\arraybackslash}p{0.255\linewidth}>{\centering\arraybackslash}p{0.105\linewidth}>{\centering\arraybackslash}p{0.095\linewidth}>{\centering\arraybackslash}p{0.115\linewidth}Y@{}}
\toprule
Variant & \shortstack{\footnotesize Macro gain\\\footnotesize (pp)} & \shortstack{\footnotesize Neg. run\\\footnotesize (\%)} & \shortstack{\footnotesize Worst combo\\\footnotesize (pp)} & Interpretation \\
\midrule
Full \method{} & $+0.118$ & $18.4$ & $-0.041$ & Complete method/\allowbreak reference \\
Effect-only$^{\dagger}$ & $+0.048$ & $21.3$ & $0.000$ & Matched-coverage diagnostic \\
No matched-null & $+0.001$ & $25.7$ & $-0.075$ & Null calibration removed \\
No coverage/MDL & $-0.436$ & $45.6$ & $-4.146$ & Coverage control removed \\
\shortstack[l]{\footnotesize No node-level filtering\\\footnotesize + provenance veto} & $+0.068$ & $28.7$ & $-0.209$ & LCB/\allowbreak survival filtering and provenance veto removed \\
MAP & $-3.9\!\times\!10^{-5}$ & $1.5$ & $-0.005$ & State averaging replaced by MAP; near no-action \\
\bottomrule
\end{tabularx}
\vspace{2pt}

\noindent\parbox{0.99\linewidth}{\footnotesize $^{\dagger}$Effect-only inherits \method's frozen changed-count budget and is not independently deployable. Node-level filtering comprises Combined-LCB screening and cluster-survival protection. Worst combo denotes the minimum combination-level mean and is distinct from a catastrophic-run count.}
\end{table}
\FloatBarrier

Table~5 compares the variants by macro gain, negative-run rate, and worst-combination performance. This experiment uses the historical frozen 15-combination/136-run registry, whereas the current broader native-interface evaluation uses the 15-combination/113-run protocol. The results are therefore used only to characterize changes in the gain--risk profile after method simplification and are not compared directly with the absolute gains in the broader evaluation. No tested simplification simultaneously preserves the mean gain, negative-run behavior, and execution constraints of complete SHR. This ablation does not isolate the causal contribution of each component.

\noindent\textbf{Auxiliary relation-source ablation.} For the source ablation, the downstream SHR procedure is fixed and only the relation used to construct the candidate residual is changed. We compared original-graph propagation, an attribute $k$-NN graph, a single-scale attribute hypergraph, the full multiscale attribute hypergraph, and a randomized hypergraph that preserved node incidence degree and hyperedge size. The randomized hypergraph preserves node incidence degree and hyperedge size while shuffling the node--hyperedge correspondence. All auxiliary sources used the same attribute preprocessing, representation alignment, and mapping to the clustering coordinate system. Refinement strength was calibrated against the same target hard-assignment change ratios, so the sources had the same target coverage states at the candidate-generation stage.

\begin{figure}[!htbp]
\centering
\includegraphics[width=0.98\linewidth]{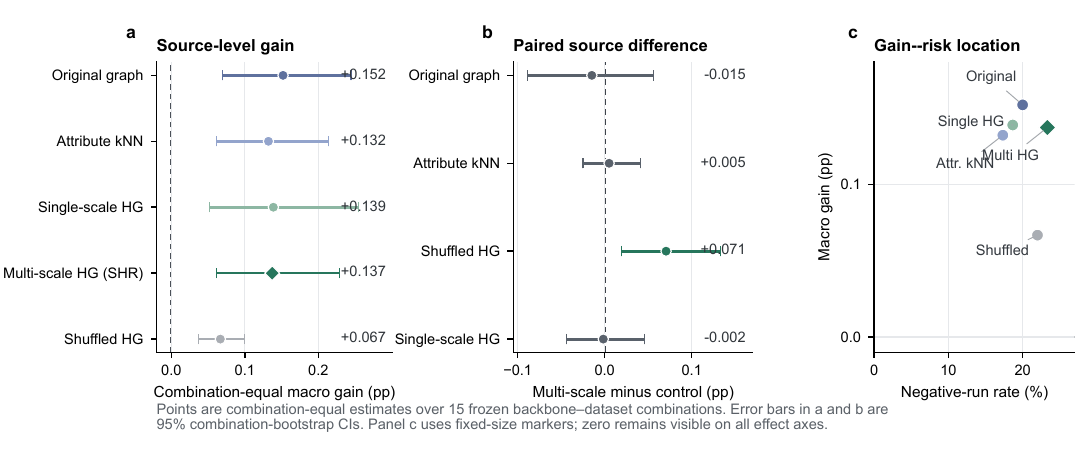}
\caption{Auxiliary relation-source ablation under the unified frozen \method{} protocol. (a) Combination-equal four-metric macro gain for five auxiliary relation sources across the 15 backbone--dataset combinations; error bars denote 95\% bootstrap CIs with the combination as the sampling unit. (b) Paired macro-gain difference between the full multiscale hypergraph source and each control; the vertical dashed line denotes zero difference. (c) Combination-equal macro gain and negative-run rate for each source. The five shuffled hypergraphs are first averaged within the same frozen run and are not treated as independent runs. Final change ratios can differ among sources, so panel b represents the overall effect of replacing the source under a common operational protocol rather than a strictly final-coverage-matched direct structural effect.}
\label{fig:hypergraph_source_ablation}
\end{figure}

Under the current frozen 15-combination/113-run protocol, the full multiscale attribute hypergraph achieved a combination-equal macro gain of $+0.137\pp$, with a 95\% CI of $[0.062,0.228]\pp$. As shown in Figure~\ref{fig:hypergraph_source_ablation}, its paired difference\footnote{Paired difference denotes the difference between the macro gain of the multiscale hypergraph source and that of the corresponding control on the same frozen units, aggregated under the prespecified combination-level convention. For the randomized hypergraph, the five randomized outcomes within the same frozen run are averaged first and are not treated as five independent runs.} relative to the randomized hypergraph was $+0.071\pp$, with a 95\% CI of $[0.019,0.133]\pp$, indicating an incremental contribution of the true node--hyperedge correspondence to the complete SHR refinement process. By contrast, the differences relative to the attribute $k$-NN graph, original-graph propagation, and single-scale attribute hypergraph were $+0.005\pp$, $-0.015\pp$, and $-0.002\pp$, with corresponding 95\% CIs of $[-0.025,0.041]\pp$, $[-0.088,0.056]\pp$, and $[-0.044,0.046]\pp$, all of which crossed zero. The current results therefore do not establish that the multiscale hypergraph is consistently superior to these ordinary relation sources. Meanwhile, the randomized hypergraph itself achieved a macro gain of $+0.067\pp$ (95\% CI, $[0.037,0.099]\pp$), whereas the multiscale hypergraph had the highest negative-run rate among the five sources, at $23.3\%$. Although the sources were calibrated to the same target change ratios during candidate generation, their final change ratios were not identical after safety projection and node screening; for example, the multiscale and randomized sources changed $0.375\%$ and $0.229\%$ of nodes, respectively. Because the final change ratios differ after projection and node filtering, these comparisons measure the operational effect of replacing the relation source rather than a strictly coverage-matched structural effect. The true node--hyperedge correspondence shows an incremental advantage over the shuffled control, while the current results do not show a consistent advantage over the ordinary graph, attribute $k$-NN, or single-scale hypergraph sources. Additional protocol details are provided in Appendix~\ref{app:source_ablation_details}.

\FloatBarrier
\subsection{Mechanism Analysis}
\label{sec:mechanism}

\begin{figure}[!htbp]
    \centering
    \includegraphics[width=0.78\linewidth]
    {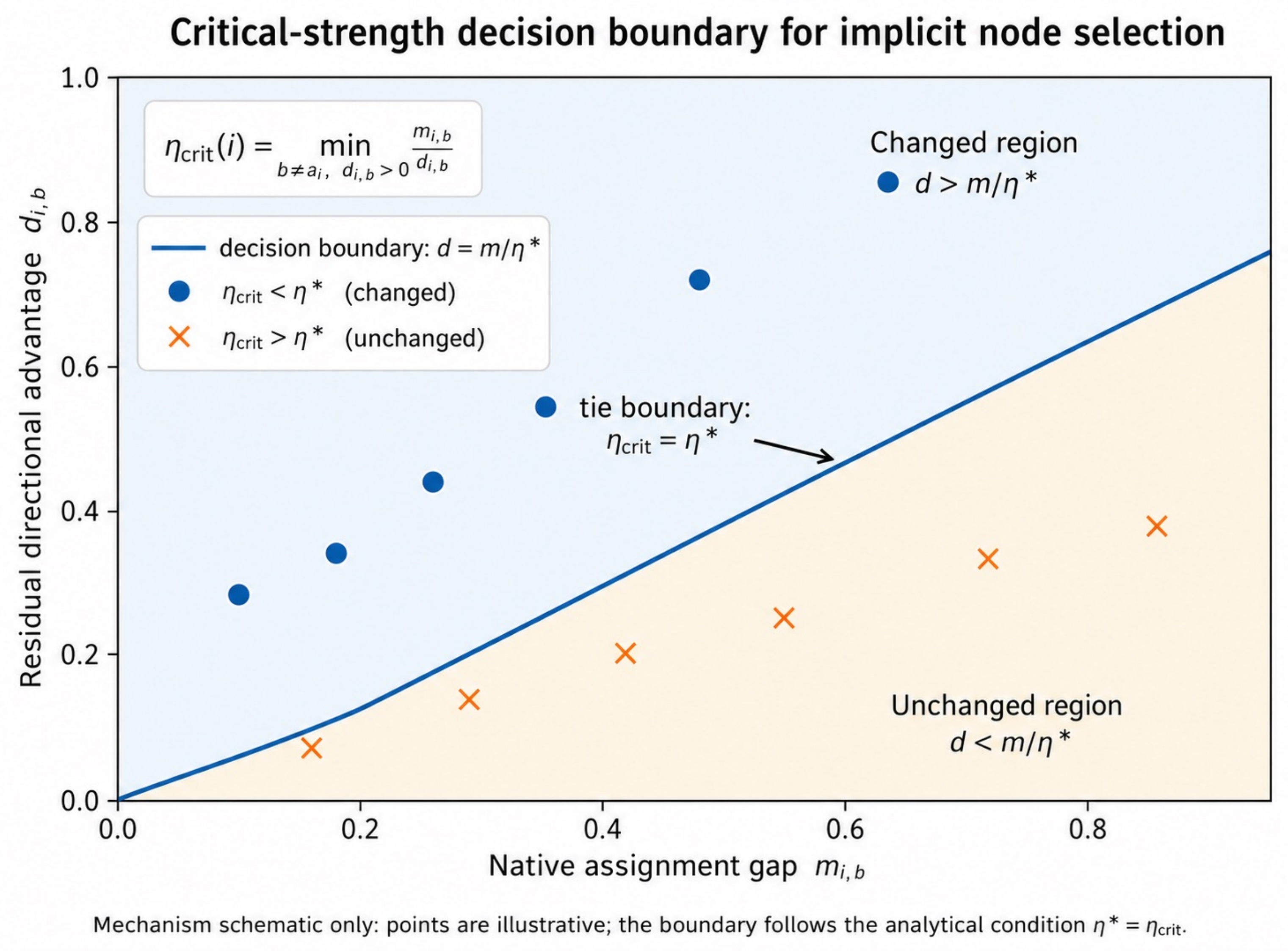}
    \caption{Geometric interpretation of critical refinement strength. For a fixed global strength $\eta^*$, the assignment boundary satisfies $d=m/\eta^*$. Nodes above the boundary have smaller $\eta_{\mathrm{crit}}=m/d$ and therefore change assignment earlier. Equality denotes tied cluster scores. The plotted nodes illustrate the analytical mechanism and do not represent an empirical distribution.}
    \label{fig:eta_crit_mechanism}
\end{figure}

\noindent\textbf{Critical refinement strength and implicit node selection.}
Section~\ref{sec:implicit_selection} shows that, even when all nodes receive the same global refinement strength, they have different critical strengths $\eta_{\mathrm{crit}}$ because their native assignment gaps and relative residual pushes differ. For a fixed global strength, the changed mask predicted by $\eta\geq\eta_{\mathrm{crit}}(i)$ exactly matched the changes obtained by direct application of the residual. We then ranked nodes by $\eta_{\mathrm{crit}}$, formed a candidate changed set at the same target change ratio, and compared it with the globally changed set obtained by direct calibration. Across all 100 frozen runs, the mean Jaccard similarity between the two sets was 0.930; when only the 89 runs with nonzero refinement were considered, it was 0.922.%
\footnote{The Jaccard coefficient is the size of the intersection of two changed-node sets divided by the size of their union; values closer to 1 indicate greater agreement. A comparison between two empty sets is defined as 1 in this analysis.}
The global residual therefore reaches nodes with smaller critical strengths first, producing an implicit ordering of candidate changes. SHR subsequently decides which of these candidate changes are retained. As shown in Figure~\ref{fig:eta_crit_mechanism}, for a fixed global strength $\eta^*$, the condition for a node to reach the assignment boundary is $\eta^*\geq\eta_{\mathrm{crit}}$, equivalently $d\geq m/\eta^*$. Equality denotes tied cluster scores, whereas a strict hard-assignment change requires the node to cross the boundary.

\FloatBarrier

\begin{figure}[tbp]
\centering
\includegraphics[width=.95\textwidth]{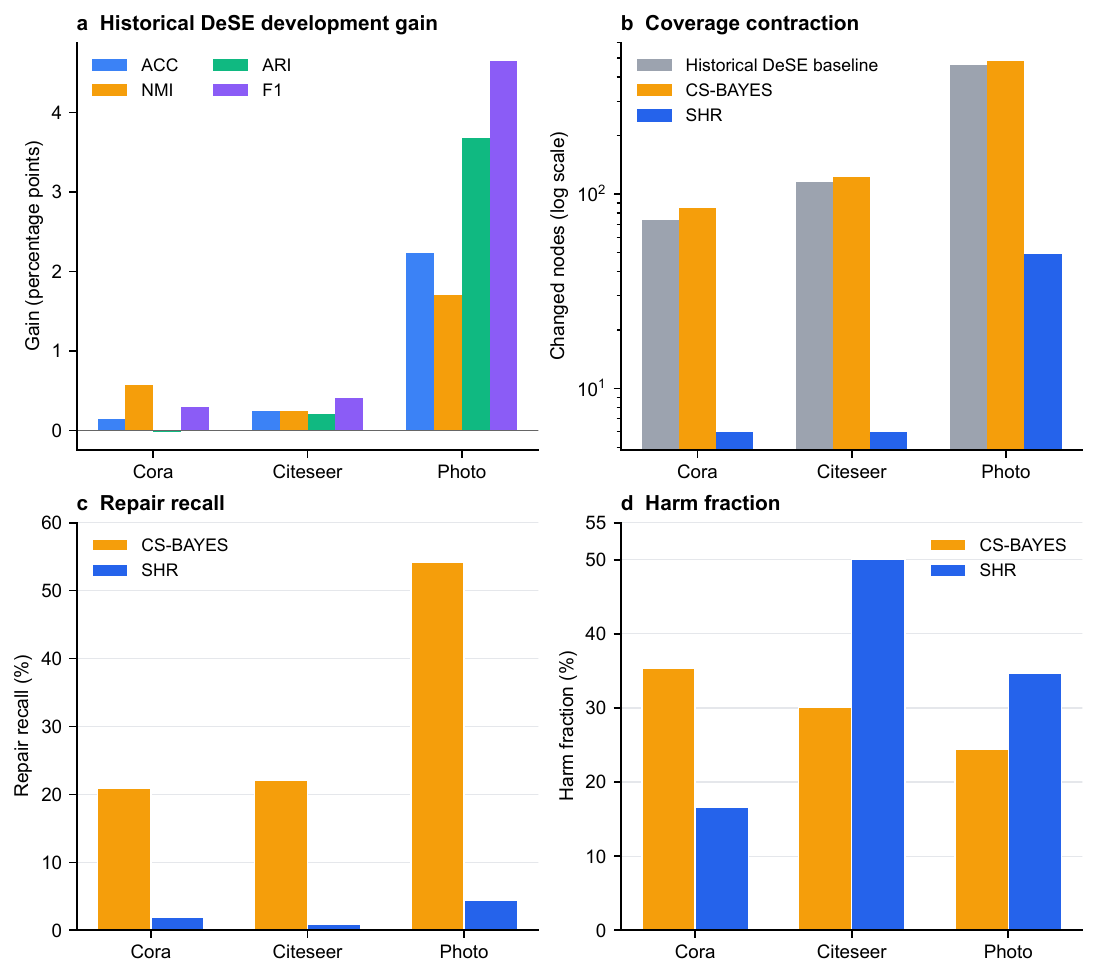}
\caption{Development-stage improvement space and coverage differences on DeSE. (a) Clustering gains of the historical DeSE development baseline over the frozen outputs on Cora, Citeseer, and Photo. (b) Changed-node counts for the historical development baseline, CS-BAYES, and SHR. (c) Repair Recall for CS-BAYES and SHR. (d) Harm Fraction among changed nodes. SHR ultimately changes substantially fewer nodes, but its node-level Harm Fraction is not always lower. Repair and harm are post hoc diagnostics computed only after method outputs are frozen and do not enter SHR decisions.}
\label{fig:dese}
\end{figure}

\FloatBarrier
\noindent\textbf{Refinement coverage and node-level outcomes.}
Critical refinement strength determines the order in which nodes enter the changed set, whereas the final coverage determines how many potential errors a method can reach. The DeSE results show that broader-coverage refinements generally achieve higher Repair Recall. SHR ultimately changes fewer nodes and therefore also has substantially lower Repair Recall. Smaller coverage, however, does not automatically produce a lower Harm Fraction: on Citeseer and Photo, the proportion of harmful changes under SHR was not lower than that under CS-BAYES. SHR is conservative mainly because it changes fewer nodes, not because the current evidence reliably identifies every beneficial refinement. On Cora, Citeseer, and Photo under the native DeSE soft-assignment interface, the historical DeSE development baseline achieved a three-dataset mean macro gain of $1.195\pp$ over the frozen outputs. Re-executing the saved historical procedure produced a maximum reproduction error of no more than $4.3\times10^{-6}$ among the reported metrics; complete results are provided in Appendix~\ref{app:dese_development}. Under the same frozen residual, the macro gains of CS-BAYES on the three datasets were $0.166$, $0.350$, and $3.136\pp$, respectively. SHR changed 6, 6, and 49 nodes, whereas the historical development baseline changed 74, 116, and 463 nodes. Figure~\ref{fig:dese} shows that the gain differences occur together with changes in intervention coverage and node-level filtering, while the candidate residual remains fixed. Strict definitions of repair, harm, and their diagnostic coordinate system are provided in Appendix~\ref{app:repair_harm_diagnostics}.

\FloatBarrier
\noindent\textbf{Identifiability of label-free evidence.}
Post hoc identifiability diagnostics further indicate that the current unlabeled selector remains a limiting factor. In the fixed candidate space, 9 of the 10 active combinations retain at least $0.10\pp$ of oracle--global headroom, while the Combined-LCB repair--harm ROC--AUC is $0.684$. Direction-stability and cross-system probes also show no consistent improvement. The current selector therefore appears more effective at limiting intervention coverage than at identifying every beneficial refinement. Full results are reported in Appendix~E.1.

\FloatBarrier
\subsection{Risk and Applicability Boundaries}
\label{sec:risk_scope}

\noindent\textbf{Gain and risk.}
The refinement strategies show different gain--risk profiles under the same frozen outputs and candidate residual. CS-BAYES had a combination-equal macro gain of $0.348\pp$, compared with $0.137\pp$ for SHR; meanwhile, SHR had 3 catastrophic runs among 113 frozen units, whereas CS-BAYES had 12, and their worst-combination macro gains were $-0.041\pp$ and $-0.993\pp$, respectively. The paired bootstrap intervals for the SHR--CS-BAYES macro-gain difference and negative-run-rate difference both crossed zero. The results show different empirical gain--risk profiles, but they do not support a statistically significant safety advantage for SHR.

\begin{figure}[!htbp]
\centering
\includegraphics[width=\textwidth]{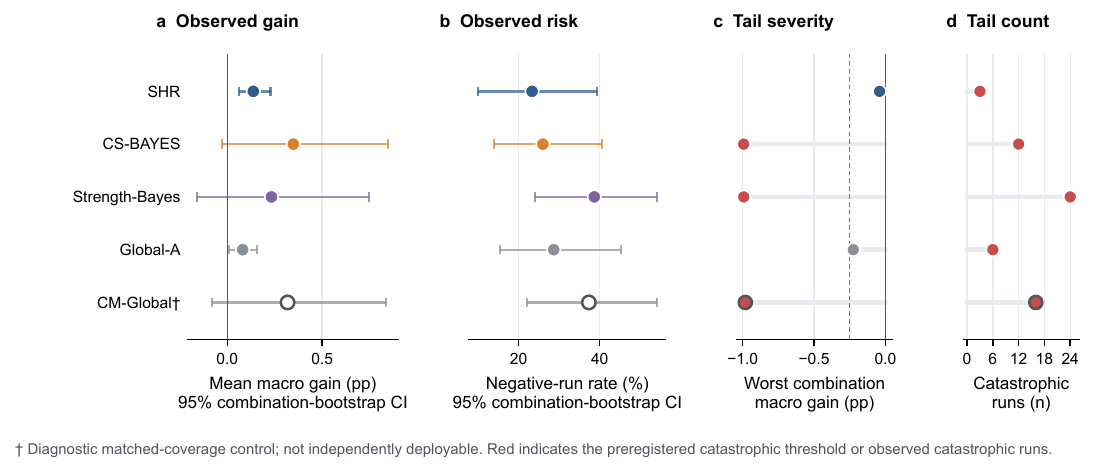}
\caption{Gain--risk comparison under a shared residual. (a) Combination-equal macro gain. (b) Combination-equal negative-run rate. (c) Worst-combination macro gain. (d) Catastrophic-run count. Error bars denote 95\% combination-bootstrap CIs. CM-Global inherits its coverage from CS-BAYES and therefore serves only as a matched-coverage diagnostic rather than an independent method.}
\label{fig:risk}
\end{figure}

\FloatBarrier
Complete gain--risk values are reported in Appendix~\ref{app:gain_risk_details}.

\noindent\textbf{Role of coverage.}
Coverage-Matched Global Residual (CM-Global) inherits the changed-node count of CS-BAYES in each run and approximately matches that coverage by adjusting global residual strength; it is therefore a mechanism diagnostic rather than an independently deployable method. CS-BAYES exceeds CM-Global by only $0.030\pp$ in combination-equal macro gain. Together with Section~\ref{sec:mechanism}, this result shows that broader coverage coincides with larger observed gains in the current candidate space, while also exposing more nodes to harmful changes. The comparison does not isolate a causal effect of coverage.

\noindent\textbf{Boundaries and applicability.}
SHR requires the frozen model to provide valid soft cluster assignments aligned with the original clustering coordinate system. When this interface condition is not met, the method can return the no-action state. For example, strict-final DeSE on Photo and Computers had fewer active clusters than the declared $K$, so the cluster-survival constraint prevented refinement. DBCD--Cora was likewise retained only as a boundary case. The preregistered $4\times5$ matrix further showed that native cluster collapse for DeSE on Cora, ACM, and UAT caused the matrix to fail its interface gate; common-suite backbone inclusion was therefore determined by five-dataset interface validity rather than SHR gain. A valid interface does not guarantee a positive SHR gain. Under the common-suite estimand, DGAC--Citeseer produced exact no-action and SynC--UAT had a macro gain of $-0.042\pp$; the latter was $-0.041\pp$ in the broader native-interface registry. The current label-free evidence is more reliable for restricting the refinement range than for consistently identifying beneficial changes across datasets. Coverage and gain have no simple monotonic relation: broader coverage can increase both repair and harm. The observed gain--risk pattern is empirical and does not provide a performance guarantee. The label-free claim applies only to the SHR refinement stage after checkpoint fixation. Complete boundary cases, interface failures, and collapse counts are reported in Appendix~\ref{app:boundary_cases}.

\FloatBarrier
\section{Conclusion}
\label{sec:conclusion}

We study label-free post-processing of frozen graph-clustering models after checkpoint fixation and propose Selective Hypergraph Refinement (SHR). SHR keeps the backbone parameters, node representations, and original graph structure fixed. An attribute hypergraph provides the candidate residual, while label-free evidence determines which assignment changes are retained. The label-free constraint applies only to the \method{} refinement stage after checkpoint fixation and does not include prior backbone training or checkpoint selection.

The controlled common suite shows that frozen clustering outputs retain a limited but measurable refinement space after training. Among 15 backbone--dataset cells, 13 had a positive mean gain, one produced exact no-action, and one was negative; the cell-equal macro gain was $0.066\pp$, while only $0.209\%$ of hard assignments changed on average. The broader five-backbone, 15-combination native-interface evaluation yielded a macro gain of $0.137\pp$ at a mean change ratio of $0.375\%$, again with substantial heterogeneity. Because only a small fraction of assignments is changed, SHR is better viewed as local post-processing of the frozen clustering output rather than reconstruction of a new global clustering result. Critical-strength analysis shows that node responses depend on both the native assignment gap and the relative residual push toward candidate clusters. Broader coverage reaches more potentially repairable nodes but also exposes more nodes to harmful changes.

The source ablation shows an incremental effect of the true node--hyperedge correspondence relative to the shuffled hypergraph under the same frozen protocol. However, the full multiscale attribute hypergraph did not consistently outperform an attribute $k$-NN graph or original-graph propagation. The results support the use of genuine auxiliary relational information, but the multiscale hypergraph does not show a consistent advantage over the alternative relation sources considered here. The current label-free evidence also remains limited in distinguishing beneficial from harmful refinements. Improving this label-free selection remains an important direction for frozen graph-clustering post-processing.

\section*{Data and code availability}

All datasets used in this study are publicly available; their sources and preprocessing details are provided in the experimental setup and Supplementary Information. The code used for SHR, result aggregation, and metric computation, together with machine-readable records supporting the primary results, will be released through a public code repository upon acceptance.

During the SHR refinement stage, node labels are used only for external evaluation and post hoc analysis after all method outputs have been frozen. They do not enter candidate refinement, state weighting, or node selection.

\section*{Generative AI Disclosure}

Generative AI tools were used extensively during manuscript preparation,
including for language editing and code-related assistance. All AI-assisted
material was reviewed and verified by the author, who takes full responsibility
for the scientific content, experimental results, references, figures, and
conclusions. Generative AI tools are not authors.

\bibliographystyle{unsrtnat}
\bibliography{references}

\clearpage
\appendix

\section{Implementation Details and Frozen Protocol}
\label{app:implementation}

\subsection{SHR Implementation Details}
\label{app:shr_implementation}

After truncated SVD and standardization of the attributes, an attribute hypergraph is constructed with $k_h=10$; the multiscale representation is constructed according to Eq.~\eqref{eq:multiscale_hyper}. The auxiliary $Q_h$ enters the context only and is not used as a direct output. We next specify the state weighting used in the formal implementation. For $r\in\mathcal S$, let $C_r=\{i:\arg\max Q^{(\eta_r)}_i\neq\arg\max Q_{0,i}\}$ and $c_r=|C_r|$. In the $m$th repeated held-out fold, let $T^{(m)}$ and $F^{(m)}$ denote the topology and attribute score matrices. The node utility and state statistic implemented in the code are
\begin{align}
u_{i,r}^{(m)}&=\tfrac12\langle Q^{(\eta_r)}_i-Q_{0,i},T_i^{(m)}\rangle
+\tfrac12\langle Q^{(\eta_r)}_i-Q_{0,i},F_i^{(m)}\rangle
-0.05\,\operatorname{JS}(Q^{(\eta_r)}_i,Q_{0,i}),\\
s_r&=\frac{1}{M}\sum_{m=1}^{M}\frac{1}{c_r}\sum_{i\in C_r}u_{i,r}^{(m)},
\end{align}
where $M=5$; when $c_r=0$, the implementation sets $s_r=0$. Each matched-null repetition permutes the rows of $\Delta$ within strata defined by the baseline cluster and confidence quintile, and recalibrates the null state to the same target hard-assignment change ratio. For the $B=19$ null statistics $s_{r,j}^{\mathrm{null}}$, we define
\begin{equation}
b_r=\sum_{j=1}^{B}\mathbf 1[s_{r,j}^{\mathrm{null}}\ge s_r],\qquad
e_r=\max\!\left(\frac{B-b_r}{1+b_r},\frac{1}{B+1}\right).
\end{equation}
The coverage complexity is computed exactly as in the code using the Beta--Binomial count prior and the MDL term:
\begin{equation}
L(c)=\log\operatorname{BetaBinomial}(c;N,1,19)-\tfrac12\log(1+c).
\end{equation}
Let $\mathcal S$ contain the seven nonzero effect states. The log weights for the no-action and nonzero states, their normalized weights, and the resulting output are
\begin{align}
\ell_0&=L(0), & \ell_r&=\log e_r+L(c_r)-\log|\mathcal S|,\\
\omega_j&=\frac{\exp(\ell_j)}{\exp(\ell_0)+\sum_{r\in\mathcal S}\exp(\ell_r)}, &
Q_{\mathrm{mix}}&=\operatorname{Normalize}\!\left(\omega_0Q_0+\sum_{r\in\mathcal S}\omega_rQ^{(\eta_r)}\right).
\end{align}
Thus, the no-action state participates directly in the averaging through $Q_0$ and its weight $\omega_0$, rather than being added after the averaging step. The projection sequentially evaluates
\begin{equation}
Q_{\tau}=\operatorname{Normalize}((1-\tau)Q_0+\tau Q_{\mathrm{mix}}),
\qquad \tau\in\{1,0.75,0.5,0.25,0\},
\end{equation}
and adopts the first state that satisfies the frozen hard-assignment-change-ratio, mean-JS, and active-cluster constraints. Positive-LCB filtering then reverts unsupported rows, and failure of the attribute-provenance condition causes the entire run to return $Q_0$. These normalized evidence weights are based solely on label-free empirical evidence and the complexity term; they are not posterior probabilities or probabilities of correctness.

\subsection{Frozen Implementation Constants}
\label{app:frozen_constants}

Table~\ref{tab:constants} summarizes the implementation constants shared by all frozen evaluations.

\begin{table}[!htbp]
\centering
\caption{Frozen implementation constants for the formal \method{} configuration.}
\label{tab:constants}
\scriptsize
\begin{tabularx}{\textwidth}{l l Y Y}
\toprule
Parameter & Value & Role & Selection rule / provenance\\
\midrule
Attribute SVD width & $\min(64,D_X-1,N-1)$ & Auxiliary attribute signal & Fixed implementation rule\\
Alignment width & $\min(32,D_{Z_0},D_{H_h})$ & Common Procrustes coordinates & Fixed implementation rule\\
Ridge $\lambda$ & $10^{-2}$ & Native-logit decoder & Frozen pre-intervention implementation\\
Hypergraph $k_h$ & 10 & Attribute kNN hyperedges & Shared across datasets/backbones\\
Effect targets & $.005,.01,.02,.03,.05,.08,.12$ & Target hard-assignment change ratios & Frozen target grid\\
$\eta$ calibration grid & $0$ plus 80 log-spaced points in $[.01,20]$ & Match effect targets & Frozen search grid\\
Evidence repeats/folds & $5/5$ & Predictive evidence & Fixed repeated-fold protocol\\
JS evidence penalty & $.05$ & Penalize assignment displacement & Frozen implementation\\
Matched-null repetitions & 19 & Empirical odds $e_r$ & Frozen permutation budget\\
Coverage prior & Beta--Binomial$(N,1,19)$ & Sparse count prior & Frozen implementation\\
MDL coefficient & $.5$ & $\log(1+c)$ complexity penalty & Frozen implementation\\
Projection $\tau$ & $1,.75,.5,.25,0$ & Conservative shrinkage & Fixed ordered path\\
Max hard-change ratio & $.20$ & Projection constraint & Frozen protocol\\
Mean JS threshold & $.10$ & Projection constraint & Frozen protocol\\
LCB coefficient & $1.645$ & One-sided node filter & Frozen 95\% coefficient\\
Provenance condition & retained attribute support mean $>0$ & Run-level accept/reject & Frozen hard-reject rule\\
\bottomrule
\end{tabularx}
\end{table}

\clearpage
\section{Evaluation Protocol and Metric Definitions}
\label{app:evaluation}

\subsection{Label-Isolation Protocol}

The label-isolation protocol separates the construction and evaluation stages. The construction stage reads only the frozen graph, attributes, representation, $Q_0$, residual, and existing label-free signals, and writes the candidate states, changed masks, coverage values, and selection outputs to disk before labels are accessed. The evaluation stage then reads the labels to compute external clustering metrics, repair/harm, oracle quantities, AUC, and tail precision. All intervention outputs are frozen before labels are accessed, and no post-hoc quantity is fed back into candidate generation, state weighting, node filtering, or the final output.

\subsection{Clustering Metrics}

For each run and prediction $\hat y$, ACC first constructs a contingency table between the ground-truth labels and predicted clusters, and then applies Hungarian assignment to maximize the matched count; the baseline and intervention mappings are obtained independently. F1 is computed as the macro average after applying the same prediction-specific mapping, with \texttt{zero\_division=0}. NMI uses arithmetic normalization, and ARI uses the adjusted Rand definition; both are invariant to permutations of cluster identifiers.

\subsection{Repair, Harm, and Oracle Diagnostics}
\label{app:repair_harm_diagnostics}

Repair/harm uses a distinct but fixed diagnostic coordinate system: a Hungarian mapping $M_0$ is fitted only to the baseline prediction, and $M_0$ is then used to assess both baseline and intervention correctness. Let
\begin{equation}
h_i=\mathbf 1[\hat y_i^*\ne \hat y_i^0],\qquad
c_i^0=\mathbf 1[M_0(\hat y_i^0)=y_i],\qquad
c_i^*=\mathbf 1[M_0(\hat y_i^*)=y_i],
\end{equation}
then
\begin{equation}
\mathrm{repair}_i=h_i(1-c_i^0)c_i^*,\quad
\mathrm{harm}_i=h_ic_i^0(1-c_i^*),\quad
\mathrm{neutral}_i=h_i-\mathrm{repair}_i-\mathrm{harm}_i.
\end{equation}
Repairable residual candidates have finite $\eta_{\mathrm{crit},i}$, a defined $b_{\mathrm{crit},i}$, $c_i^0=0$, and $M_0(b_{\mathrm{crit},i})=y_i$. The oracle is allowed to evaluate only the residual-specified $b_{\mathrm{crit},i}$ and cannot choose a different target class. Repair Recall is the number of actual repairs divided by this fixed candidate pool and is recorded as NA when the denominator is empty. Harm Fraction is the number of harmful changes divided by the changed-node count; when the changed-node count is 0, the implementation records it as 0.

Accepted-harm nodes are changed nodes retained by SHR for which $\mathrm{harm}_i=1$; that is, they were correct under the fixed baseline mapping $M_0$ but became incorrect after SHR refinement. Missed-repair nodes are drawn from the same fixed eligible residual-candidate pool; they were not ultimately selected by SHR, although the residual-specified $b_{\mathrm{crit},i}$ would change a baseline error into a correct assignment under $M_0$. Both categories are determined with labels only after method outputs are frozen and do not enter SHR node selection.

\clearpage
\section{Datasets, Backbones and Provenance}
\label{app:provenance}

\subsection{Dataset Statistics and Preprocessing}
\label{app:datasets}

Table~\ref{tab:datasets} reports the actual inputs included in the current, development, or boundary registries. Edge counts are the numbers of undirected edges after the loader removes self-loops, deduplicates edges, and symmetrizes the graph. \method{} does not override each backbone's feature preprocessing; instead, it reconstructs the auxiliary signal from the frozen input attributes by applying truncated SVD followed by standardization, and uses $k_h=10$ for the attribute kNN hypergraph. For graph evidence and the residual context, self-loops are added temporarily only within the propagation operator. When the same dataset is used by different backbones, backbone-specific upstream preprocessing is inherited from the corresponding released implementation.

\begin{table}[!htbp]
\centering
\caption{Dataset statistics and sources corresponding to the actual loaders and artifacts. DBCD--Cora is a separate interface boundary; its declared $K=19$ is inherited from this frozen interface and does not replace the standard Cora value of $K=7$.}
\label{tab:datasets}
\scriptsize
\resizebox{\textwidth}{!}{%
\begin{tabular}{lrrrrll}
\toprule
Dataset & Nodes & Edges & Features & $K$ & Backbone(s) & Input / preprocessing source\\
\midrule
Cora & 2708 & 5278 & 1433 & 7 & DeSE, DGAC, SynC & PyG Planetoid canonical undirected graph\\
Citeseer & 3327 & 4552 & 3703 & 6 & DeSE, DGAC, SynC & PyG Planetoid canonical undirected graph\\
Photo & 7650 & 119081 & 745 & 8 & DeSE & PyG Amazon canonical undirected graph\\
Computers & 13752 & 245861 & 767 & 10 & DeSE legacy/boundary & PyG Amazon canonical undirected graph\\
ACM & 3025 & 13128 & 1870 & 3 & HALO, SynC & Released NPY; diagonal removed; symmetrized\\
UAT & 1190 & 13599 & 239 & 4 & HALO, SynC & Released NPY; diagonal removed; symmetrized\\
EAT & 399 & 5993 & 203 & 4 & HALO & Released HALO NPY; diagonal removed; symmetrized\\
BAT & 131 & 1003 & 81 & 4 & HALO & Released HALO NPY; diagonal removed; symmetrized\\
DBLP & 4057 & 3528 & 334 & 4 & SynC & Released SynC NPY; diagonal removed; symmetrized\\
Adam & 3660 & 14270 & 10 & 8 & DGM & Frozen strict feature/edge artifacts; symmetrized\\
DBCD--Cora & 2708 & 5278 & 1433 & 19 & DBCD & Planetoid raw reconstruction; interface boundary\\
\bottomrule
\end{tabular}%
}
\end{table}

\FloatBarrier
\subsection{Backbone and Checkpoint Provenance}

Upstream training and checkpoint-selection protocols follow the corresponding backbone implementations and are reported separately in Table~\ref{tab:provenance}. Once $Z_0$ and $Q_0$ are fixed, every SHR intervention decision follows the same label-free protocol. A paper citation identifies the backbone model, whereas local code and artifact provenance establish the checkpoint rule; the DeSE label-aware checkpoint claim is therefore not attributed to the paper alone.

\begin{table}[!htbp]
\centering
\caption{Upstream provenance separated from SHR intervention. ``GT'' denotes ground-truth labels; ``Val'' denotes validation labels. UNKNOWN evidence is not silently interpreted as label-free.}
\label{tab:provenance}
\scriptsize
\resizebox{\textwidth}{!}{%
\begin{tabular}{lllllll}
\toprule
Backbone & Datasets & Checkpoint rule & GT upstream & Val upstream & Native $Q_0$ & SHR labels\\
\midrule
DeSE~\citep{zhang2025dese} & Cora/Citeseer/Photo & released paper-best, ACC-selected & YES & NO & YES & NO\\
HALO~\citep{zhu2026halo} & ACM/UAT/EAT/BAT & fixed final epoch & NO & NO & YES & NO\\
DGAC~\citep{xie2025dgac} & Cora/Citeseer & fixed final epoch & NO & NO & YES & NO\\
SynC~\citep{ding2026sync} & Cora/Citeseer/ACM/UAT/DBLP & fixed final epochs & NO & NO & YES & NO\\
DGM~\citep{liu2026dgm} & Adam & AE epoch 200; DGM epoch 500 & NO & NO & YES & NO\\
DBCD~\citep{wang2026dbcd} & Cora & fixed final epoch & NO & NO & YES & NO\\
\bottomrule
\end{tabular}%
}
\end{table}
\FloatBarrier
The numerical DBCD final epoch and exact released-code commits for DeSE, DGAC and DGM were not preserved in the current frozen provenance records. This does not alter the locally recorded selection rule, but limits commit-level reproducibility. Additional code provenance will be provided with the anonymized code repository.

\subsection{Boundary and Interface Cases}
\label{app:boundary_cases}

\noindent\textbf{Strict-final DeSE Photo/Computers.}
For these strict-final anchors, the number of active clusters is smaller than the declared $K$, so the cluster-survival constraint returns no-action. These runs are not included in the primary effectiveness analysis; the resulting zero intervention gain does not imply zero baseline clustering performance.

\noindent\textbf{DBCD--Cora.}
The native assignment and active-cluster interface do not satisfy the primary protocol requirements, so this combination is reported only as an interface boundary case.

\noindent\textbf{Computers legacy readout.}
The development study uses a frozen DeSE embedding followed by an unsupervised KMeans readout. This is a legacy, non-native diagnostic interface: it is reported separately, excluded from the native DeSE mean, and not used to support a four-dataset same-interface claim. Replaying the historical DeSE development baseline on the native Cora/Citeseer/Photo interfaces yields a maximum absolute metric error of $4.3\times10^{-6}$; the Computers readout does not fully reproduce the paper-reported Computers row.

\noindent\textbf{Negative boundary behavior.}
SynC--UAT remains a negative primary case with a macro gain of $-0.041\pp$, illustrating that satisfying the interface conditions does not guarantee positive refinement gain. DGAC--BAT provides an additional negative outcome under aggressive coverage and is reported separately from the primary 15-combination analysis.

\noindent\textbf{Failed preregistered $4\times5$ matrix and common-suite derivation.}
The unified-matrix audit completed all 20 smoke cells and all 100 formal runs for DeSE, HALO, DGAC, and SynC on Cora, Citeseer, ACM, UAT, and DBLP. For DeSE on Cora, ACM, and UAT, at least three seeds had fewer native active clusters than the declared $K$, so the complete $4\times5$ matrix failed the preregistered native-interface validity gate. The subsequent common suite retains only HALO, DGAC, and SynC, each of which has a valid native $Q_0$ on all five datasets. This subset is determined by interface validity rather than SHR gain. Table~\ref{tab:common_suite_interface_audit} retains the complete collapse counts and inclusion decisions.

\begin{table}[!htbp]
\centering
\caption{Technical audit of the attempted preregistered $4\times5$ matrix. All 20 smoke cells and 100 formal runs completed. DeSE failed the native-interface validity gate on three datasets because at least three of five runs had fewer active native clusters than the declared $K$. The subsequent $3\times5$ common suite is therefore interface-validity-driven rather than performance-selected.}
\label{tab:common_suite_interface_audit}
\resizebox{\linewidth}{!}{%
\begin{tabular}{llrrll}
\toprule
Backbone & Dataset & Completed & Collapsed & Valid interface & Common-suite decision \\
\midrule
DeSE & Cora & 5 & 5 & No & Excluded: interface invalid \\
DeSE & Citeseer & 5 & 2 & Yes & Not retained: backbone incomplete across suite \\
DeSE & ACM & 5 & 4 & No & Excluded: interface invalid \\
DeSE & UAT & 5 & 5 & No & Excluded: interface invalid \\
DeSE & DBLP & 5 & 2 & Yes & Not retained: backbone incomplete across suite \\
HALO & All five & 25 & 0 & Yes & Included by interface validity \\
DGAC & All five & 25 & 0 & Yes & Included by interface validity \\
SynC & All five & 25 & 0 & Yes & Included by interface validity \\
\bottomrule
\end{tabular}%
}
\end{table}

\FloatBarrier
\thispagestyle{plain}
\section{Additional Results and Robustness}
\label{app:additional_results}

\subsection{Common-Suite Statistics and Post-Processing Cost}
\label{app:common_suite}

Table~\ref{tab:common_suite_cell_stats} reports complete cell-level statistics for the common suite. All five paired seeds are retained in every cell; the cell-level bootstrap intervals are descriptive for these small samples and are not used for individual-cell significance claims. Table~\ref{tab:common_suite_intervention_runtime_cells} reports the changed-node count, change ratio, changed nodes per 1,000, and post-processing runtime. Timing begins after $A$, $X$, $Z_0$, and $Q_0$ have been prepared and ends when formal SHR returns $Q^*$. Each unit receives one warm-up and three timed repetitions, and backbone training is excluded. All 75 timed outputs exactly match their corresponding frozen SHR outputs.

\begin{table}[!htbp]
\centering
\caption{Cell-level common-suite macro-gain statistics. Values are percentage points. Intervals are descriptive percentile intervals from 10,000 paired seed-level bootstrap resamples ($n=5$ per cell; seed 4102026). Counts give positive/exact-zero/negative frozen runs.}
\label{tab:common_suite_cell_stats}
\resizebox{\linewidth}{!}{%
\begin{tabular}{llrrrrr}
\toprule
Backbone & Dataset & Mean $\pm$ SD & Median & 95\% CI & $+ / 0 / -$ & $n$ \\
\midrule
HALO & Cora & +0.050 $\pm$ 0.015 & +0.054 & [+0.036, +0.058] & 5 / 0 / 0 & 5 \\
HALO & Citeseer & +0.083 $\pm$ 0.074 & +0.055 & [+0.033, +0.148] & 5 / 0 / 0 & 5 \\
HALO & ACM & +0.170 $\pm$ 0.107 & +0.114 & [+0.095, +0.264] & 5 / 0 / 0 & 5 \\
HALO & UAT & +0.181 $\pm$ 0.235 & +0.263 & [-0.002, +0.362] & 3 / 1 / 1 & 5 \\
HALO & DBLP & +0.004 $\pm$ 0.010 & 0.000 & [0.000, +0.013] & 1 / 4 / 0 & 5 \\
DGAC & Cora & +0.115 $\pm$ 0.078 & +0.124 & [+0.054, +0.170] & 4 / 1 / 0 & 5 \\
DGAC & Citeseer & 0.000 $\pm$ 0.000 & 0.000 & [0.000, 0.000] & 0 / 5 / 0 & 5 \\
DGAC & ACM & +0.008 $\pm$ 0.011 & 0.000 & [0.000, +0.017] & 2 / 3 / 0 & 5 \\
DGAC & UAT & +0.096 $\pm$ 0.137 & +0.062 & [+0.012, +0.216] & 3 / 2 / 0 & 5 \\
DGAC & DBLP & +0.011 $\pm$ 0.015 & +0.017 & [0.000, +0.022] & 3 / 1 / 1 & 5 \\
SynC & Cora & +0.224 $\pm$ 0.097 & +0.184 & [+0.150, +0.302] & 5 / 0 / 0 & 5 \\
SynC & Citeseer & +0.002 $\pm$ 0.029 & 0.000 & [-0.021, +0.027] & 1 / 3 / 1 & 5 \\
SynC & ACM & +0.033 $\pm$ 0.065 & +0.011 & [-0.012, +0.088] & 3 / 0 / 2 & 5 \\
SynC & UAT & -0.042 $\pm$ 0.139 & -0.004 & [-0.157, +0.055] & 2 / 0 / 3 & 5 \\
SynC & DBLP & +0.063 $\pm$ 0.121 & +0.046 & [-0.022, +0.163] & 4 / 0 / 1 & 5 \\
\bottomrule
\end{tabular}%
}
\end{table}

\FloatBarrier
\begin{table}[!htbp]
\centering
\caption{Detailed intervention scale and runtime for the 15 common-suite cells. Change statistics summarize five paired frozen runs; runtime is the median of the five per-unit median timings.}
\label{tab:common_suite_intervention_runtime_cells}
\resizebox{\linewidth}{!}{%
\begin{tabular}{llrrrrr}
\toprule
Backbone & Dataset & Changed nodes & Change (\%) & Range (\%) & Changed/1000 & Time (s) \\
\midrule
HALO & Cora & 9.4 & 0.347 & 0.222--0.591 & 3.47 & 1.74 \\
HALO & Citeseer & 5.4 & 0.162 & 0.090--0.301 & 1.62 & 2.15 \\
HALO & ACM & 4.8 & 0.159 & 0.066--0.331 & 1.59 & 1.53 \\
HALO & UAT & 4.8 & 0.403 & 0.000--0.588 & 4.03 & 0.92 \\
HALO & DBLP & 0.2 & 0.005 & 0.000--0.025 & 0.05 & 1.98 \\
DGAC & Cora & 9.2 & 0.340 & 0.000--0.554 & 3.40 & 1.76 \\
DGAC & Citeseer & 0.0 & 0.000 & 0.000--0.000 & 0.00 & 2.13 \\
DGAC & ACM & 2.4 & 0.079 & 0.000--0.298 & 0.79 & 1.50 \\
DGAC & UAT & 1.8 & 0.151 & 0.000--0.504 & 1.51 & 0.74 \\
DGAC & DBLP & 3.0 & 0.074 & 0.000--0.173 & 0.74 & 2.04 \\
SynC & Cora & 17.0 & 0.628 & 0.517--0.739 & 6.28 & 1.68 \\
SynC & Citeseer & 1.0 & 0.030 & 0.000--0.120 & 0.30 & 2.04 \\
SynC & ACM & 4.8 & 0.159 & 0.066--0.231 & 1.59 & 1.46 \\
SynC & UAT & 4.6 & 0.387 & 0.168--0.588 & 3.87 & 1.00 \\
SynC & DBLP & 8.6 & 0.212 & 0.099--0.345 & 2.12 & 1.90 \\
\bottomrule
\end{tabular}%
}
\end{table}

\FloatBarrier

\subsection{Comparator Details}
\label{app:comparators}

Table~\ref{tab:comparators} summarizes the coverage, strength, state-averaging scheme, and experimental role of each comparison method under the shared candidate residual; the corresponding state spaces and computational rules are specified below.

\begin{table}[!htbp]
\centering
\caption{Comparison methods under the shared candidate residual. CM-Global inherits its refinement coverage from CS-BAYES and is therefore used only as a matched-coverage diagnostic.}
\label{tab:comparators}
\scriptsize
\begin{tabularx}{\textwidth}{l Y Y Y c Y}
\toprule
Method & Coverage & Strength & State averaging & Standalone & Role \\
\midrule
Global-A & Global & Fixed $\eta=.1$ & None & Yes & Fixed-strength baseline \\
Strength-Bayes & $\pi=1$ if active & Eight $\eta$ states & Posterior average + no-action & Yes & Strength uncertainty \\
CS-BAYES & Nine positive $\pi$ states & Eight $\eta$ states & Joint $\pi/\eta/z_i$ average & Yes & Coverage--strength comparator \\
\method{} & Effect-state proposal + node filter & Effect-calibrated & Evidence-weighted + no-action & Yes & Proposed coverage-conservative method \\
CM-Global & Inherits CS-BAYES count & Coverage-matched search & None & No & Diagnostic only \\
\bottomrule
\end{tabularx}
\end{table}

Global-A is defined as $Q^{A}=\operatorname{softmax}(\log(Q_0+10^{-8})+0.1\Delta)$ and uses neither a posterior nor the output of another method. For Strength-Bayes, the nonzero states fix $\pi=1$ and use $\eta\in\{0.7,1,1.5,2,3,5,7,10\}$; for CS-BAYES, the positive-coverage states are $\pi\in\{.01,.03,.05,.1,.2,.4,.6,.8,1\}$, with $\gamma=0$ fixed. Both share the no-action prior $p_0=.1$ and the discrete strength prior
\begin{equation}
\log p(\eta_e)= -\tfrac12(\eta_e/5)^2-\operatorname{logsumexp}_{e'}[-\tfrac12(\eta_{e'}/5)^2].
\end{equation}
Let $\ell_{i,e}$ denote the local label-free log-evidence for node $i$ at strength $e$. The run-level generalized evidence is $T_N\sum_i\log[(1-\pi)+\pi\exp(\ell_{i,e})]$, where $T_N=\min(1,512/N)$. Strength-Bayes fixes the positive states at $\pi=1$; CS-BAYES performs posterior-style averaging over $(\pi,\eta)$ and the conditional node keep/action probabilities rather than using the posterior mode. Both then evaluate the maximum change rate, mean JS, and active-cluster survival along the fixed shrinkage path $\tau\in\{1,.75,.5,.25,0\}$.

CM-Global first reads the changed-node count $k$ produced by CS-BAYES for the same run, and then selects, from $\eta=0$ and $[0.01,20]$ sampled on an 80-point geometric grid, the global residual state with the smallest lexicographic key $(|\#\mathrm{changed}-k|,\overline{\mathrm{JS}})$. CM-Global therefore cannot determine coverage independently and serves only as a same-coverage diagnostic.

\FloatBarrier
\subsection{DeSE Development Results}
\label{app:dese_development}

The historical DeSE development baseline records the development-stage improvement space observed when post-processing frozen DeSE outputs. On Cora, Citeseer, and Photo under the native soft-assignment interface, its three-dataset mean macro gain over the frozen outputs is $1.195\pp$. Re-executing the saved historical procedure yields a maximum absolute reproduction error of $4.3\times10^{-6}$ among the reported metrics. This development gain is improvement space observed in the historical development procedure, not a theoretical performance upper bound. Complete values are listed in Table~\ref{tab:dese_development}. The Computers row uses the legacy non-native readout described in Appendix~\ref{app:boundary_cases} and is excluded from the mean over the three native-interface datasets.

\begin{table}[!htbp]
\centering
\caption{DeSE development results on native assignments and the legacy Computers readout.}
\label{tab:dese_development}
\scriptsize
\resizebox{\textwidth}{!}{%
\begin{tabular}{llrrrrr}
\toprule
\multicolumn{7}{l}{\textbf{Historical DeSE development baseline on native frozen assignments}}\\
Dataset & Role & $\Delta$ACC & $\Delta$NMI & $\Delta$ARI & $\Delta$F1 & Changed\\
\midrule
Cora & development & .148 & .566 & -.018 & .293 & 74\\
Citeseer & development & .240 & .249 & .199 & .404 & 116\\
Photo & development & 2.235 & 1.703 & 3.679 & 4.645 & 463\\
Native mean & development & .874 & .839 & 1.287 & 1.780 & --\\
\midrule
\multicolumn{7}{l}{\textbf{Legacy non-native readout (excluded from native mean)}}\\
Computers & frozen embedding + KMeans & 1.825 & .819 & 1.454 & .624 & 1315\\
\bottomrule
\end{tabular}%
}
\end{table}

\FloatBarrier
\subsection{Complete SHR Results}
\label{app:complete_results}

Table~\ref{tab:complete_shr_results} reports the combination-level absolute results for the fixed 15-combination primary registry. It is aggregated directly from the existing frozen records for 113 units by taking the arithmetic mean over frozen units within each backbone--dataset combination. Strict-final degenerate anchors, the DBCD interface boundary, the Computers legacy readout, and additional aggressive-control diagnostics are excluded from this table.

\begin{table}[!htbp]
\centering
\caption{Complete absolute results for SHR across the 15 primary backbone--dataset combinations (Part I: frozen-baseline and final SHR metrics). Absolute metrics are reported on the $[0,1]$ scale, and each row is the descriptive arithmetic mean over the corresponding frozen units. Depending on the backbone, units are seeds, folds, or a single frozen checkpoint; therefore, the DeSE rows with $n=1$ do not estimate run-to-run uncertainty. The frozen baseline denotes the locally frozen output rather than a directly transcribed paper-reported score. The upstream selection of the DeSE checkpoints used labels; the label-free constraint applies only to the SHR stage after checkpoint fixation.}
\label{tab:complete_shr_results}
\scriptsize
\resizebox{\textwidth}{!}{%
\begin{tabular}{llr rrrr rrrr}
\toprule
& & & \multicolumn{4}{c}{Frozen baseline} & \multicolumn{4}{c}{SHR final}\\
\cmidrule(lr){4-7}\cmidrule(lr){8-11}
Backbone & Dataset & $n$ & ACC & NMI & ARI & F1 & ACC & NMI & ARI & F1\\
\midrule
DeSE & Cora & 1 & 0.7518 & 0.5703 & 0.5270 & 0.7099 & 0.7526 & 0.5716 & 0.5275 & 0.7108 \\
DeSE & Citeseer & 1 & 0.6889 & 0.4398 & 0.4474 & 0.6399 & 0.6886 & 0.4394 & 0.4468 & 0.6397 \\
DeSE & Photo & 1 & 0.8046 & 0.6901 & 0.6241 & 0.7318 & 0.8055 & 0.6917 & 0.6252 & 0.7378 \\
DGAC & Cora & 5 & 0.6393 & 0.5339 & 0.4381 & 0.6344 & 0.6405 & 0.5351 & 0.4397 & 0.6350 \\
DGAC & Citeseer & 5 & 0.6909 & 0.4350 & 0.4450 & 0.6444 & 0.6909 & 0.4350 & 0.4450 & 0.6444 \\
DGM & Adam & 10 & 0.7518 & 0.7589 & 0.6851 & 0.7002 & 0.7517 & 0.7589 & 0.6851 & 0.7001 \\
HALO & ACM & 10 & 0.8125 & 0.5070 & 0.5466 & 0.8112 & 0.8137 & 0.5090 & 0.5488 & 0.8124 \\
HALO & BAT & 10 & 0.5237 & 0.2911 & 0.2184 & 0.4962 & 0.5290 & 0.2992 & 0.2254 & 0.5003 \\
HALO & EAT & 10 & 0.4739 & 0.1926 & 0.1690 & 0.4617 & 0.4757 & 0.1981 & 0.1728 & 0.4633 \\
HALO & UAT & 10 & 0.4861 & 0.2221 & 0.1861 & 0.4552 & 0.4882 & 0.2243 & 0.1883 & 0.4574 \\
SynC & ACM & 10 & 0.9148 & 0.7023 & 0.7628 & 0.9151 & 0.9152 & 0.7030 & 0.7639 & 0.9155 \\
SynC & Citeseer & 10 & 0.6951 & 0.4415 & 0.4525 & 0.6449 & 0.6950 & 0.4415 & 0.4525 & 0.6447 \\
SynC & Cora & 10 & 0.6613 & 0.4978 & 0.4292 & 0.6494 & 0.6635 & 0.5008 & 0.4325 & 0.6514 \\
SynC & DBLP & 10 & 0.8042 & 0.5054 & 0.5595 & 0.7961 & 0.8046 & 0.5065 & 0.5604 & 0.7964 \\
SynC & UAT & 10 & 0.5561 & 0.2601 & 0.2481 & 0.5489 & 0.5557 & 0.2596 & 0.2480 & 0.5484 \\
\bottomrule
\end{tabular}%
}
\end{table}

\begin{table}[!htbp]
\ContinuedFloat
\centering
\caption[]{Complete absolute results for SHR (continued: metric gains relative to the frozen baseline, macro gain, and hard-assignment change ratio). Metric gains and macro gain are reported in percentage points; Change is reported as a percentage. The valid no-action result for DGAC--Citeseer and all negative combinations are retained exactly as frozen. Machine-readable records for every frozen unit and unrounded combination-level values will be released with the code repository.}
\scriptsize
\begin{tabular}{llr rrrrr r}
\toprule
Backbone & Dataset & $n$ & $\Delta$ACC & $\Delta$NMI & $\Delta$ARI & $\Delta$F1 & Macro & Change (\%)\\
\midrule
DeSE & Cora & 1 & +0.074 & +0.126 & +0.047 & +0.093 & +0.085 & 0.222 \\
DeSE & Citeseer & 1 & -0.030 & -0.037 & -0.062 & -0.023 & -0.038 & 0.180 \\
DeSE & Photo & 1 & +0.092 & +0.158 & +0.106 & +0.597 & +0.238 & 0.641 \\
DGAC & Cora & 5 & +0.118 & +0.120 & +0.161 & +0.060 & +0.115 & 0.340 \\
DGAC & Citeseer & 5 & 0.000 & 0.000 & 0.000 & 0.000 & 0.000 & 0.000 \\
DGM & Adam & 10 & -0.008 & +0.000 & +0.000 & -0.009 & -0.004 & 0.030 \\
HALO & ACM & 10 & +0.116 & +0.200 & +0.222 & +0.119 & +0.164 & 0.162 \\
HALO & BAT & 10 & +0.534 & +0.807 & +0.699 & +0.411 & +0.613 & 1.221 \\
HALO & EAT & 10 & +0.175 & +0.552 & +0.376 & +0.159 & +0.316 & 1.103 \\
HALO & UAT & 10 & +0.210 & +0.214 & +0.214 & +0.224 & +0.215 & 0.395 \\
SynC & ACM & 10 & +0.043 & +0.070 & +0.109 & +0.042 & +0.066 & 0.139 \\
SynC & Citeseer & 10 & -0.009 & +0.008 & +0.002 & -0.012 & -0.003 & 0.042 \\
SynC & Cora & 10 & +0.218 & +0.292 & +0.330 & +0.206 & +0.261 & 0.654 \\
SynC & DBLP & 10 & +0.042 & +0.117 & +0.087 & +0.039 & +0.071 & 0.197 \\
SynC & UAT & 10 & -0.034 & -0.055 & -0.016 & -0.059 & -0.041 & 0.303 \\
\bottomrule
\end{tabular}
\end{table}

\FloatBarrier
\subsection{Aggregation and Backbone Sensitivity}
\label{app:aggregation}

Table~\ref{tab:backboneequal} reports the two-stage aggregation: datasets are first averaged within each backbone, and the backbone-level means are then weighted equally; combination-equal remains the primary statistic.
\begin{table}[!htbp]
\centering
\caption{Equal-backbone sensitivity (gains in percentage points). Summary rows average the five backbone-level means equally.}
\label{tab:backboneequal}
\small
\begin{tabular}{lrrrrr}
\toprule
Method & $\Delta$ACC & $\Delta$NMI & $\Delta$ARI & $\Delta$F1 & Macro\\
\midrule
Global-A & .023 & .098 & .077 & .047 & .061\\
Strength-Bayes & .133 & .343 & .398 & .135 & .252\\
CS-BAYES & .161 & .542 & .515 & .132 & .337\\
SHR & .081 & .134 & .118 & .103 & .109\\
\bottomrule
\end{tabular}
\end{table}
For macro gain, SHR$-$CS-BAYES is $-0.228\pp$, with a 95\% backbone-bootstrap CI of $[-0.681,0.118]$. For negative-run rate, the difference is $-2.13$ percentage points, with a 95\% CI of $[-26.8,22.0]$. SHR itself has an equal-backbone macro gain of $0.109\pp$ (exploratory 95\% bootstrap CI $[0.031,0.222]\pp$). With only five bootstrap units, these are exploratory sensitivity intervals rather than confirmatory evidence.

\FloatBarrier
\subsection{Detailed Gain--Risk Comparison}
\label{app:gain_risk_details}

Table~\ref{tab:risk_summary} preserves the complete numerical gain--risk comparison moved from the main text. Standalone methods use the combination-equal risk estimand; CM-Global inherits the coverage of CS-BAYES and is only a matched-coverage mechanism diagnostic.

\begin{table}[!htbp]
\centering
\caption{Gain--risk summary under a shared residual. Standalone methods use combination-equal risk; CM-Global is only a matched-coverage mechanism diagnostic. Catastrophic counts are reported over $N=113$ frozen paired runs per method.}
\label{tab:risk_summary}
\scriptsize
\resizebox{\textwidth}{!}{%
\begin{tabular}{llrrrrr}
\toprule
\multicolumn{7}{l}{\textbf{Standalone interventions, 15-combination audit}}\\
Method & Macro gain & Eq.-comb. neg. & Pooled neg. & Neg. combos & Catastrophic & Worst combo\\
\midrule
Global-A & .081 & 28.67\% & 24.78\% & 4/15 & 6/113 & -.225\\
Strength-Bayes & .233 & 38.67\% & 39.82\% & 7/15 & 24/113 & -.993\\
CS-BAYES & .348 & 26.00\% & 30.09\% & 4/15 & 12/113 & -.993\\
SHR & .137 & 23.33\% & 23.01\% & 4/15 & 3/113 & -.041\\
\midrule
\multicolumn{7}{l}{\textbf{Panel C: Coverage-matched mechanism diagnostic (not standalone)}}\\
CM-Global & .318 & 37.33\% & 37.17\% & 5/15 & 16/113 & -.982\\
CS-BAYES & .348 & 26.00\% & 30.09\% & 4/15 & 12/113 & -.993\\
CS $-$ CM & +.030 & -- & -- & -- & -- & --\\
\bottomrule
\end{tabular}%
}
\end{table}

\FloatBarrier
\subsection{Additional Statistical Comparison with CS-BAYES}
\label{app:cs_bayes_stats}

Across the 15 combination-level paired means, the macro-gain difference between SHR and CS-BAYES yields a two-sided Wilcoxon signed-rank statistic of $46.0$ ($p=0.454$) and a two-sided exact sign-test $p=0.607$. For the within-combination negative-run rates, the Wilcoxon statistic is $24.0$ ($p=0.421$), and the exact sign-test gives $p=1.000$. These auxiliary tests are consistent with the paired bootstrap results in the main text: the current data support neither a universal advantage for either method nor a description of SHR as statistically safer.

\FloatBarrier
\section{Additional Diagnostics and Ablation}
\label{app:diagnostics}

\subsection{Additional Identifiability Diagnostics}
\label{app:identifiability}

\begin{figure}[!htbp]
\centering
\includegraphics[width=0.92\textwidth]{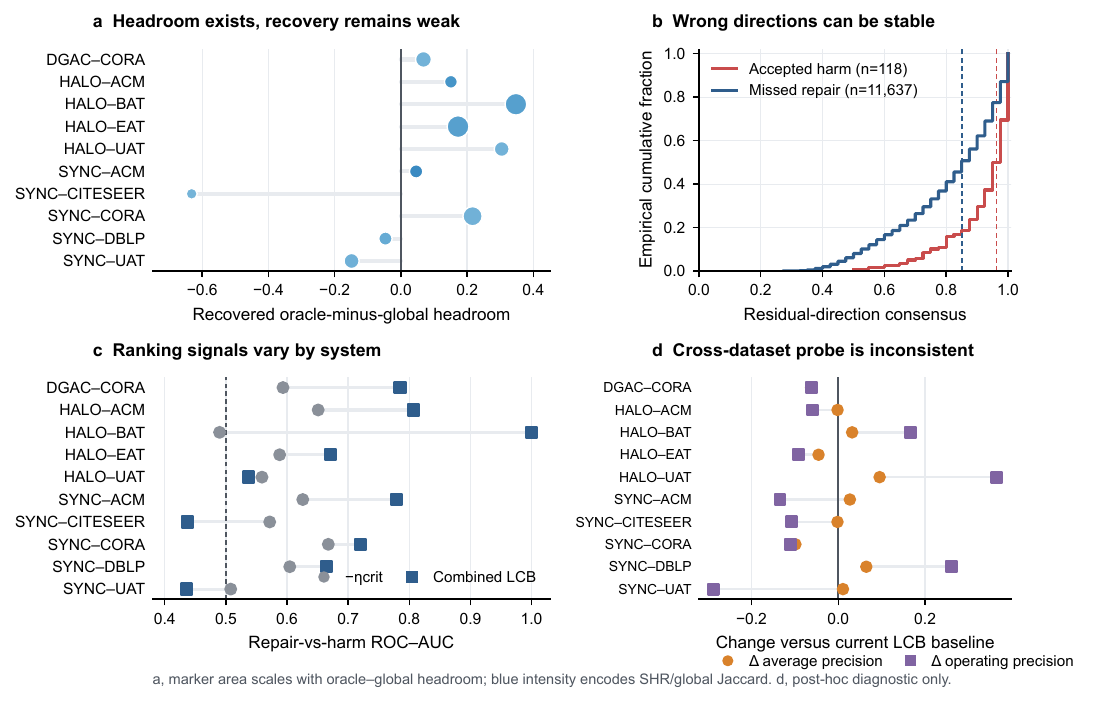}
\caption{Post hoc diagnosis of label-free selection capability. (a) Additional improvement space of the oracle over the global refinement result and its recovery by SHR for the included combinations. (b) Residual-direction consensus for nodes that were accepted but ultimately became incorrect and for unaccepted nodes that could have been repaired. (c) Repair--harm ROC--AUC for $-\eta_{\mathrm{crit}}$ and Combined-LCB. (d) Changes relative to the current LCB baseline in cross-dataset diagnosis. Repair, harm, oracle, and AUC are computed with labels only after method outputs are frozen and do not enter SHR decisions.}
\label{fig:identity}
\end{figure}

\paragraph{Diagnostic candidate set and oracle.}
This diagnostic includes only backbone--dataset combinations with at least one frozen unit whose final SHR changed-node count is nonzero. The resulting 10 combinations are DGAC--Cora; HALO--ACM, HALO--BAT, HALO--EAT, and HALO--UAT; and SynC--ACM, SynC--Citeseer, SynC--Cora, SynC--DBLP, and SynC--UAT. DGAC--Citeseer is excluded from this subset because all five frozen units return no-action. The candidate universe consists of eligible residual candidates with finite $\eta_{\mathrm{crit}}$ not exceeding the frozen upper bound of the $\eta$ search, and each target class is fixed to the residual-specified $b_{\mathrm{crit}}$. Within each run, the oracle selects $k$ nodes, where $k$ is the final SHR hard-assignment change count. The corresponding global result selects the first $k$ candidates ordered by $\eta_{\mathrm{crit}}$ and then node ID, again using $b_{\mathrm{crit}}$. The macro oracle first forms a baseline label--cluster contingency table and groups candidates by true-label row, baseline cluster, and $b_{\mathrm{crit}}$ target, with candidates within each group ordered by $\eta_{\mathrm{crit}}$ and node ID. It then adds the feasible move that maximizes the four-metric macro score at each step and performs up to 20 deterministic one-swap passes, accepting only swaps that improve the macro score by more than $10^{-12}$. This greedy-plus-one-swap solution is a feasible lower bound within the fixed candidate directions and is not guaranteed to be the combinatorial optimum. It uses labels only after output freezing, generates no new direction, does not enter SHR, and is not deployable. Let $g_{\mathrm{oracle}}$, $g_{\mathrm{global}}$, and $g_{\mathrm{SHR}}$ denote their respective macro gains. Figure~\ref{fig:identity} uses
\[
H_{\mathrm{oracle-global}}=g_{\mathrm{oracle}}-g_{\mathrm{global}},\qquad
\mathrm{Recovery}=\frac{g_{\mathrm{SHR}}-g_{\mathrm{global}}}{g_{\mathrm{oracle}}-g_{\mathrm{global}}},
\]
with Recovery recorded as NA when the denominator is nonpositive. Nine of the 10 included combinations have combination-level mean headroom of at least $0.10\pp$.

\paragraph{Combined-LCB and repair--harm AUC.}
The node-level Combined-LCB used in Section~\ref{sec:selective_execution} and the frozen score re-audited here share the same label-free definition. For each of five evidence repeats, target-direction topology and attribute supports are combined as $u_i^{(t)}=\tfrac12(u_{i,\mathrm{topo}}^{(t)}+u_{i,\mathrm{attr}}^{(t)})$, followed by the lower confidence bound $\overline u_i-1.645s_i/\sqrt{5}$. To align the score direction with the residual-fixed repair/harm target, the reported AUC of $0.684$ uses the frozen current-candidate LCB and retains only eligible decisive candidates whose stored target equals $b_{\mathrm{crit}}$. Repair and harm labels follow the fixed baseline mapping in Appendix~\ref{app:repair_harm_diagnostics}, and neutral nodes are excluded from AUC. ROC--AUC is computed within a frozen unit only when at least two observations, both repair and harm classes, and a nonconstant finite score are available; otherwise it is NA. Valid frozen-unit AUCs are first averaged within each backbone--dataset combination, and the resulting values are averaged equally over the 10 included combinations, yielding $0.684$.

\paragraph{Residual-direction consensus.}
For every frozen run, four types of mild label-free perturbation are generated: 5\% edge dropout, 5\% feature masking, feature noise with standard deviation equal to 0.05 times the global feature standard deviation, and 5\% hypergraph-neighbor membership perturbation. Each type has 10 fixed repetitions, for $T=40$ perturbations. The auxiliary residual is reconstructed for every perturbation, and the earliest target state $b_i^{(t)}$ is recorded; the absence of a finite switching strength is represented as a distinct no-switch state. Node-level directional consistency is
\[
\mathrm{direction\_consensus}_i=\max_{s\in\{-1,0,\ldots,K-1\}}\frac{1}{40}\sum_{t=1}^{40}\mathbf 1[b_i^{(t)}=s].
\]
Accepted-harm and missed-repair nodes are pooled according to Appendix~\ref{app:repair_harm_diagnostics}. Their pooled node-level medians are $0.962$ ($n=118$) and $0.850$ ($n=11{,}637$), respectively. Here, $n$ denotes diagnostic nodes rather than independent runs; within-node and within-run dependence means that these descriptive distributions are not independent-sample inference.

\paragraph{Cross-system diagnostics.}
The supervised cross-system probes use only repair and harm candidates, with repair coded as 1 and harm as 0; neutral nodes are excluded. Leave-one-dataset-out (LODO) holds out one complete backbone--dataset combination and trains on the other nine combinations, whereas leave-one-backbone-out (LOBO) holds out one complete backbone (DGAC, HALO, or SynC). The full XGBoost probe uses 28 frozen label-free features: baseline logit margin; residual directional advantage; $\eta_{\mathrm{crit}}$; Combined-LCB for the current, $b_{\mathrm{crit}}$, and action targets; topology target support; attribute target support; their support gap and source agreement; topology action support; attribute action support; candidate and actual-SHR assignment JS; baseline entropy; baseline maximum confidence; cluster size; feature-kNN target agreement; graph-neighbor target agreement; direction consensus; original-target survival; the coefficient of variation and finite fraction of $\eta_{\mathrm{crit}}$; residual-direction margin; and the direction-consensus scores under edge dropout, feature masking, feature noise, and hypergraph perturbation. Missing values receive training-set median imputation, and training weights are the normalized inverse sample counts within each dataset--class group. XGBoost is fixed at 300 trees, depth 3, learning rate $0.05$, row and column subsampling of $0.8$, $\lambda=1$, and random seed $20260807$, with no tuning. Average precision is computed over all held-out repair/harm nodes. Operating precision ranks candidates within each held-out run, selects the top $k$, and divides the repair count by that run's observed SHR changed count $k$ before averaging over runs. The separate single-signal tail analysis retains repair, harm, and neutral candidates and evaluates operating-$k$, 0.1\%, 0.25\%, 0.5\%, 1\%, 2\%, and top-5/10/20 cutoffs; neutral candidates remain in the Precision@k denominator. Relative to the current Combined-LCB baseline, LODO XGBoost has mean $\Delta$AP of approximately $+0.002$ and mean $\Delta$ operating precision of approximately $-0.006$. Only 3 of 10 holdouts satisfy the preregistered joint-win rule, and LOBO results are likewise unstable. All supervised probes are post hoc diagnostics performed after method outputs are frozen. They are not part of unsupervised SHR and must not be interpreted as unsupervised generalization performance.

Figure~\ref{fig:identity} summarizes the oracle-headroom, residual-direction-consensus, repair--harm AUC, and cross-system probe results reported above.

\FloatBarrier
\subsection{Additional Source-Ablation Protocol Details}
\label{app:source_ablation_details}

The auxiliary-source experiment fixes $Q_0$, state calibration, matched-null, coverage constraints, LCB, and provenance execution, while replacing only the relational source used to construct the residual. The comparisons include original-graph propagation, an attribute-based cosine $10$-NN graph, a single-scale attribute hypergraph, the full multiscale attribute hypergraph, and a randomized (shuffled) hypergraph that preserves node incidence degree and hyperedge size. The five shuffled hypergraphs are averaged within the same frozen run and are not treated as independent runs. All sources use the same target coverage-state calibration protocol, but projection and node filtering can produce different final change ratios. The differences between sources therefore represent the overall operational effect of replacing the information source under the same operational protocol, rather than a direct structural effect under strictly matched final coverage.

Across the 10 frozen runs of DGM--Adam, the prespecified common-dimension rule gives $d_c=26$ for the single-scale source and $d_c=32$ for all other sources. The comparison between the multiscale and single-scale sources is therefore only a descriptive scale diagnostic and cannot be interpreted as a pure scale effect under a strictly dimension-matched condition; this dimensional difference was not adjusted according to the evaluation results.

\subsection{Method Distillation and Simplification}
\label{app:distillation}

\begin{figure}[!tbp]
\centering
\includegraphics[width=\textwidth]{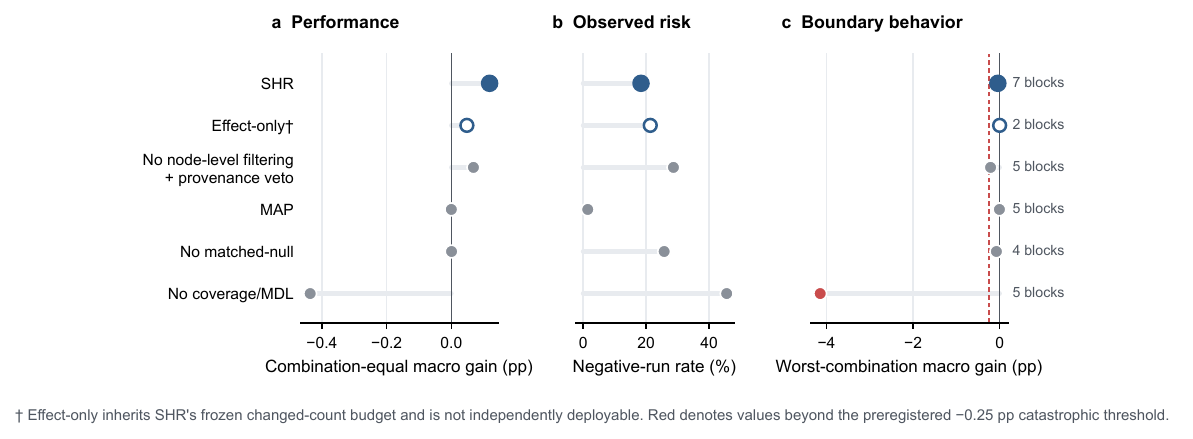}
\caption{Ablation profile of the frozen method-distillation registry. (a) Macro gain with equal weighting of backbone--dataset combinations; (b) run-weighted negative-run rate; (c) worst-combination macro gain and the catastrophic threshold defined as $-0.25\pp$, with the number of logical blocks shown on the right. No node-level filtering + provenance veto removes both Combined-LCB/cluster-survival screening and the attribute-provenance veto. Effect-only inherits the frozen changed-count budget of complete SHR and cannot be deployed independently. This figure compares the observed gain, negative-run behavior, and execution complexity of the simplified configurations and does not redefine the primary mean.}
\label{fig:distill}
\end{figure}

Distillation uses the frozen historical 15-combination/136-run registry, in which the complete-SHR macro gain is $0.118\pp$; the corresponding macro gain in the current primary 15-combination/113-run registry is $0.137\pp$. These values have different scopes and must not be combined. Effect-only retains $40.6\%$ of the observed gain of the complete configuration, but inherits its frozen changed-count budget and cannot be deployed independently. Among the simplified variants tested, No coverage/MDL, No matched-null, No node-level filtering + provenance veto, MAP, and Effect-only fail to preserve simultaneously the observed gain characteristics, negative-run behavior, and execution constraints of the complete \method. Figure~\ref{fig:distill} presents a representative performance--risk--complexity profile.

\noindent\textbf{No node-level filtering + provenance veto.} This variant retains candidate-state construction, evidence assessment, state weighting, and global safety projection, and denotes by $Q_{\mathrm{safe}}$ the candidate result that passes the global change-rate, mean-JS, and active-cluster constraints. Unlike complete SHR, it outputs $Q_{\mathrm{safe}}$ directly and omits two final layers of selective control. First, it does not apply Combined-LCB screening to the actually changed nodes or the associated cluster-survival protection. Second, it does not apply run-level hard rejection based on the mean attribute support of retained nodes. Its macro gain, negative-run rate, and worst-combination result are $+0.068\pp$, $28.7\%$, and $-0.209\pp$, respectively, compared with $+0.118\pp$, $18.4\%$, and $-0.041\pp$ for complete SHR. Under this historical protocol, the observed profile therefore combines lower mean gain, more negative runs, and a worse tail outcome; however, this ablation alone does not establish an independent causal role for either control layer.

\suppressfloats[t]

\end{document}